\documentclass[journal]{IEEEtran}
\usepackage{amsmath,amsfonts}
\usepackage{multirow}
\usepackage{algorithmic}
\usepackage{algorithm}
\usepackage{array}
\usepackage[caption=false,font=normalsize,labelfont=sf,textfont=sf]{subfig}
\usepackage{textcomp}
\usepackage{stfloats}
\usepackage{url}
\usepackage{verbatim}
\usepackage{graphicx}
\usepackage{cite}
\usepackage{color, xcolor}
\usepackage{doi}
\begin{document}

\title{Double Domain Guided Real-Time Low-Light Image Enhancement for Ultra-High-Definition Transportation Surveillance}

\author{Jingxiang Qu, Ryan Wen Liu, \textit{Member, IEEE}, Yuan Gao, Yu Guo,  

Fenghua Zhu, \textit{Senior Member, IEEE}, and Fei-Yue Wang, \textit{Fellow, IEEE}

\thanks{This work was supported by the National Key Research and Development Program of China (Grant No.: 2022YFB4300300), and the National Natural Science Foundation of China (Grant No.: 52271365).}
\thanks{Jingxiang Qu, Ryan Wen Liu, Yuan Gao, and Yu Guo are with the School of Navigation, and also with the State Key Laboratory of Maritime Technology and Safety, Wuhan University of Technology, Wuhan 430063, China (e-mail: \{qujx, wenliu, yuangao, yuguo\}@whut.edu.cn).}%
\thanks{Fenghua Zhu and Fei-Yue Wang are with the State Key Laboratory for Management and Control of Complex Systems, Institute of Automation, Chinese Academy of Sciences, Beijing 100190, China (e-mail: \{fenghua.zhu, feiyue.wang\}@ia.ac.cn).}
% \thanks{Wenqi Ren is with the School of Cyber Science and Technology, Shenzhen Campus, Sun Yat-sen University, Shenzhen 518107, China (e-mail: rwq.renwenqi@gmail.com).}

% \thanks{*The corresponding author is Ryan Wen Liu.}
}

\maketitle

\begin{abstract}
    Real-time transportation surveillance is an essential part of the intelligent transportation system (ITS). However, images captured under low-light conditions often suffer the poor visibility with types of degradation, such as noise interference and vague edge features, etc. With the development of imaging devices, the quality of the visual surveillance data is continually increasing, like 2K and 4K, which has more strict requirements on the efficiency of image processing. To satisfy the requirements on both enhancement quality and computational speed, this paper proposes a double domain guided real-time low-light image enhancement network (DDNet) for ultra-high-definition (UHD) transportation surveillance. Specifically, we design an encoder-decoder structure as the main architecture of the learning network. In particular, the enhancement processing is divided into two subtasks (i.e., color enhancement and gradient enhancement) via the proposed coarse enhancement module (CEM) and LoG-based gradient enhancement module (GEM), which are embedded in the encoder-decoder structure. It enables the network to enhance the color and edge features simultaneously. Through the decomposition and reconstruction on both color and gradient domains, our DDNet can restore the detailed feature information concealed by the darkness with better visual quality and efficiency. The evaluation experiments on standard and transportation-related datasets demonstrate that our DDNet provides superior enhancement quality and efficiency compared with the state-of-the-art methods. Besides, the object detection and scene segmentation experiments indicate the practical benefits for higher-level image analysis under low-light environments in ITS. 
    %Therefore, our DDNet can successfully satisfy the requirements of visual quality and efficiency in real-time UHD transportation surveillance. 
    The source code is available at \url{https://github.com/QuJX/DDNet}.
\end{abstract}


\section{Introduction}\label{sec:Introduction}
    \IEEEPARstart{W}{ith} the rapid growth of intelligent transportation system (ITS), more and more visual sensors are employed for transportation surveillance. However, when the imaging device is under low-light environments, the acquired images always suffer poor sharpness, low contrast, and undesirable noise \cite{jiang2022unsupervised}. The poor imaging quality makes it difficult to see the captured scenes clearly and brings great challenges to higher-level image analysis, such as object detection \cite{liang2022edge, ge2021yolox, qin2022id} and scene segmentation \cite{gao2021mscfnet, zhang2022trans4trans, weng2022deep}. Even though some imaging devices attempt to enlighten the darkness with extra artificial light such as infrared and ultraviolet flashes \cite{krishnan2009dark}, the cost and the poor quality are the main limitations. Therefore, an effective low-light image enhancement method is necessary for nocturnal transportation surveillance. Moreover, with the development of imaging and parallel computational devices, the resolution of the captured visual surveillance data is continually increasing, from the standard definition (SD, 480p, 720p), the high definition (HD, 1080p), to the ultra-high definition (UHD, 4K). The corresponding image processing algorithm has also been widely investigated under multiple transportation scenes, e.g., parking lot \cite{shen2022parkpredict+}, waterway \cite{kujawski2019concept}, and airport surveillance \cite{wang2022improved}, etc. The trade-off between visibility enhancement and computational complexity is a major problem to be solved in current transportation applications \cite{lin2022uhd}.
    \begin{figure}[t]
    	\centering
    	\setlength{\abovecaptionskip}{0.1cm}
    	\includegraphics[width=1.0\linewidth]{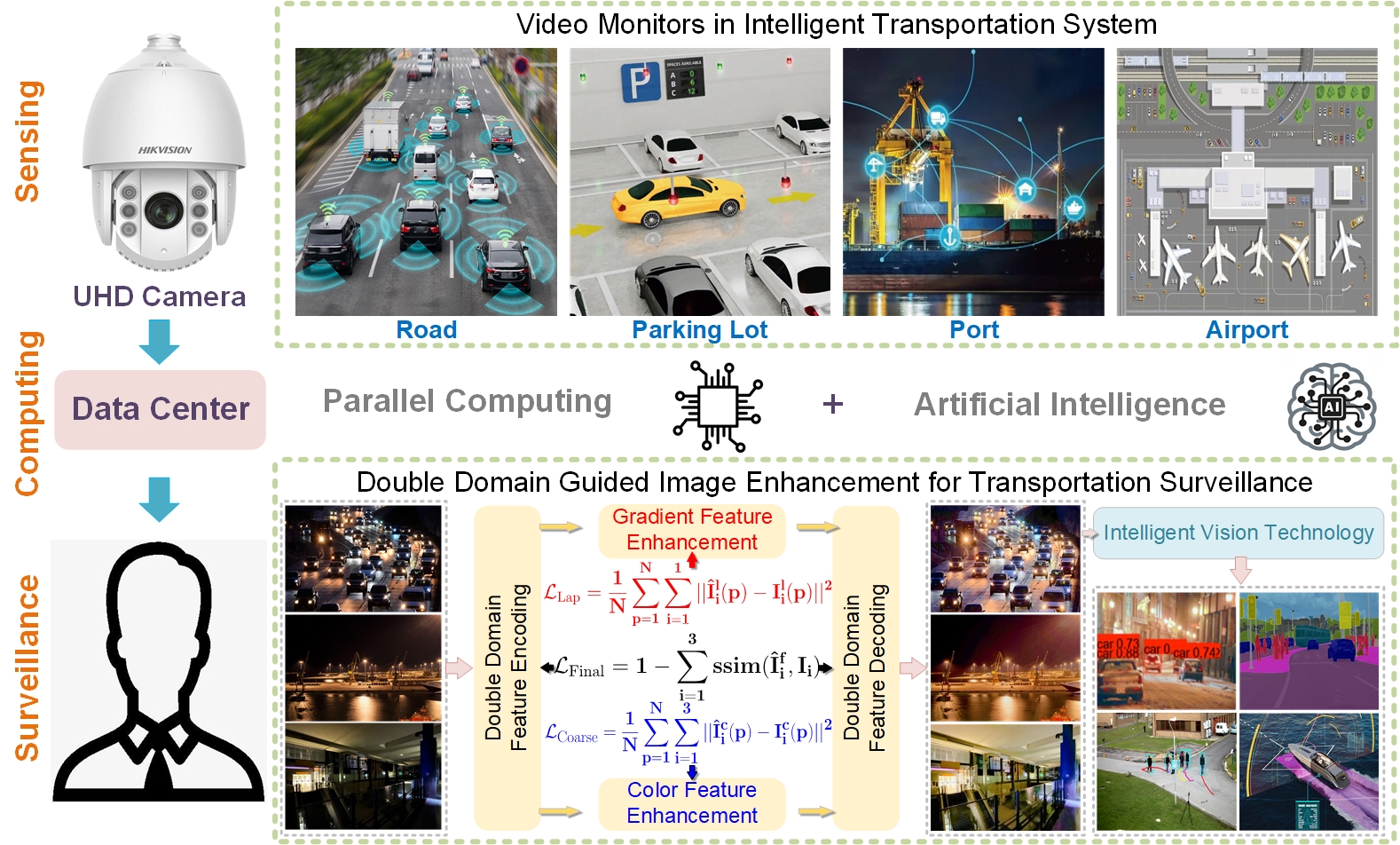}
    	\caption{The illustration of our DDNet for real-time low-light transportation surveillance under different practical scenes.}
    	\label{Applications}
    \end{figure}

\subsection{Motivation}
    The real-time transportation surveillance has two main requirements for low-light image enhancement: effectiveness and efficiency. Specifically, the main targets of transportation surveillance are vehicles \cite{zhang2021night}, pedestrians \cite{yang2020part}, vessels \cite{li2020highly}, etc. It is thus necessary to enlighten the darkness effectively with better noise suppression and feature preservation. For traditional low-light enhancement methods, the illumination is mainly improved by enhancing the contrast globally (e.g., histogram equalization (HE) \cite{HE}), which only improve visual perception without effective noise suppression. Compared with traditional methods, learning methods are robust to the noise due to the strong learning ability of deep neural networks, which could also improve the computational efficiency due to the acceleration of GPU. However, in transportation scenes, the edge feature is rarely considered in previous low-light image enhancement methods \cite{dhara2021exposedness}, which is especially important for higher-level image analysis like vehicle detection \cite{zhang2021night}, pedestrian detection \cite{wang2022multiple}, and scene segmentation \cite{yang2021ndnet}.
    \begin{figure}[t]
    	\centering
    	\setlength{\abovecaptionskip}{0.1cm}
    	\includegraphics[width=1.0\linewidth]{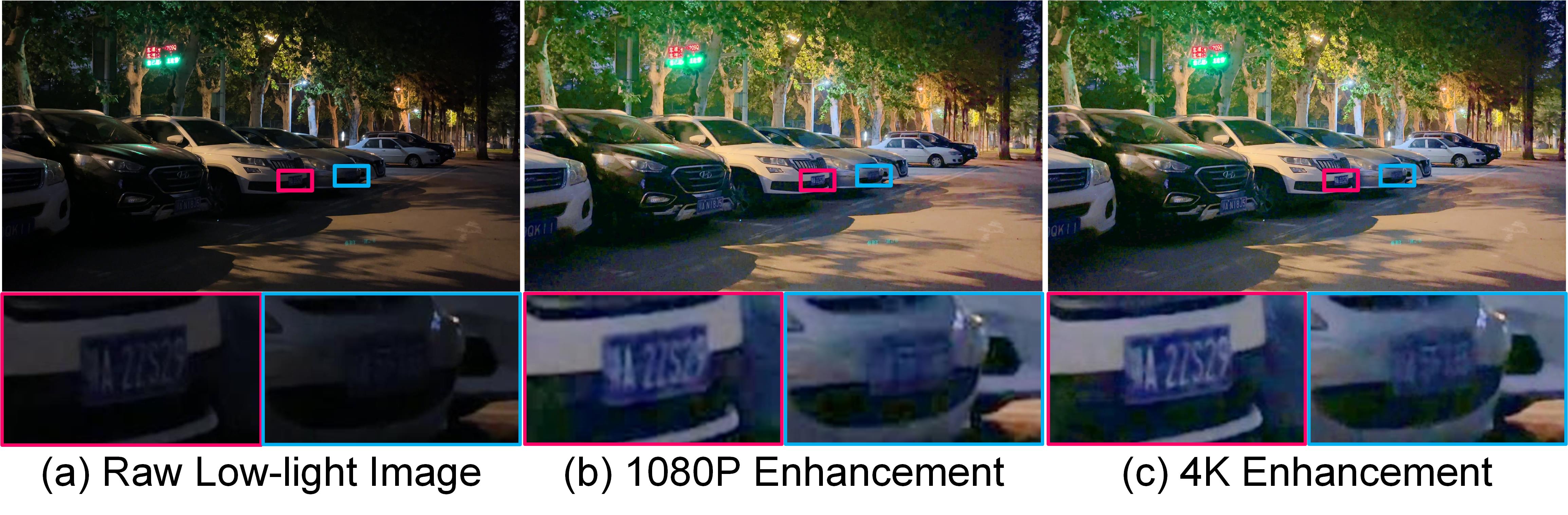}
    	\caption{The comparison between the low-light enhancement results on UHD images in transportation surveillance. From left to right: (a) raw 4K low-light image, (b) enhanced result after resizing the image to 1080P, and (c) 4K image enhancement. It is obvious that the resizing operation causes significant detail loss on UHD images.}
    	\label{4K_1080P_comparison}
    \end{figure}
    In practical applications, the frame rate of most transportation surveillance cameras is less than 30 FPS \cite{xiong2020vehicle}, which is thus the basic efficiency requirement of real-time image processing methods. However, most previous low-light image enhancement methods can not satisfy this requirement \cite{kumar2020efficient}. Therefore, in most cases, the UHD images will be firstly resized to smaller scales for lower computational complexity. It is doubtless that the image resizing has severe degeneration on the image quality. As shown in Fig. \ref{4K_1080P_comparison}, the resizing operation causes significant detail loss, making the blur of vehicle license plates. Many methods have achieved real-time processing on UHD images, like Zero \cite{guo2020zero}, SCI \cite{ma2022toward}, and UHDFour \cite{li2023embedding}, but the results are unsatisfactory in ITS scenes.
    To achieve effective real-time low-light image enhancement in UHD transportation surveillance, we propose a double domain guided network (DDNet). It achieves superior noise suppression and brightness enhancement with enhancing the feature map on the color and gradient domains simultaneously. The experiments on running time have demonstrated the efficiency of the implementation on UHD images. Furthermore, the object detection and scene segmentation experiments indicate the practical improvement for higher-level image analysis. In general, this paper provides an effective and efficient method to improve the transportation surveillance under low-light environments.
\subsection{Contributions}
    In this paper, we propose a real-time low-light image enhancement network for UHD transportation surveillance, which achieves competitive enhancement quality and computational efficiency. The main contributions of the proposed method can be summarized as follows:
    
    \begin{itemize}
    	\item We propose a double domain guided low-light image enhancement network (DDNet), aided by Laplacian of Gaussian (LoG)-based gradient information. It effectively improves the image quality captured under low-light conditions with keeping most details on both color and gradient domains.
    	\item We design the LoG-based gradient enhancement module (GEM) and the coarse enhancement module (CEM) embedded in the encoder-decoder structure, which enhances the color and gradient domain features effectively. Besides, a joint loss function is proposed to constrain the enhancement of different domains separately.
    	\item The quantitative and qualitative evaluation experiments compared with the state-of-the-arts are conducted on standard and transportation-related datasets. Experimental results show that our DDNet significantly improves the enhancement performance. Besides, the running time satisfies the requirements of real-time UHD transportation surveillance. The object detection and scene segmentation experiments indicate the improvement of our DDNet for higher-level visual tasks in ITS.

    \end{itemize}

    The rest of this paper is organized as follows. The recent studies on low-light image enhancement are reviewed in Section \ref{Related Work}. In Section \ref{Proposed Method}, We introduce the details of our DDNet. Numerous experiments on standard and transportation-related datasets have been implemented to evaluate the enhancement performance and practical benefits for transportation surveillance in Section \ref{Experiments}. Conclusion and future perspectives are finally given in Section \ref{Conclusions and Discussion}.

\section{Related Work}\label{Related Work}
    In this section, we briefly introduce the previous low-light image enhancement methods (i.e., traditional and learning methods) and their applications in ITS.

\subsection{Traditional Methods}
    The traditional methods employ some mathematical models to enhance the low-light images. Histogram equalization (HE) \cite{HE} flattens the histogram and expands the dynamic range of intensity to improve the brightness of the image. However, it is challenging to discriminate the noise and clear information with HE-based methods. Excessive noise corrupts the histogram distribution, making it harder to get reliable information from low-light backgrounds. Retinex theory \cite{retinex} and related methods \cite{improving1,improving3, cai2017joint} decompose the low-light image into the reflectance and illumination components to get the underlying normal-light image. To make a better balance between the brightness enhancement and noise suppression. However, Retinex-based methods have two major drawbacks. First, insufficient brightness enhancement in complex scenes results in unqualified enhanced images. Besides, they have difficulty in balancing noise suppression and edge feature preservation. Ying \textit{et al}. \cite{ying2017new,ying2017bio} suggested a camera response model to improve the effect of low-light image enhancement. Dong \cite{dong2010fast} and DeHz \cite{jiang2013night} enhanced the low lightness based on the atmospherical scattering model. SRRP \cite{9743313} kept the smoothness of the original illumination to achieve qualified image enhancement. However, they failed to simultaneously achieve satisfactory detail preservation, illumination enhancement, and computational efficiency for real-time UHD transportation surveillance.
    \begin{figure}[t]
        \centering
        \setlength{\abovecaptionskip}{0.1cm}
        \includegraphics[width=1.0\linewidth]{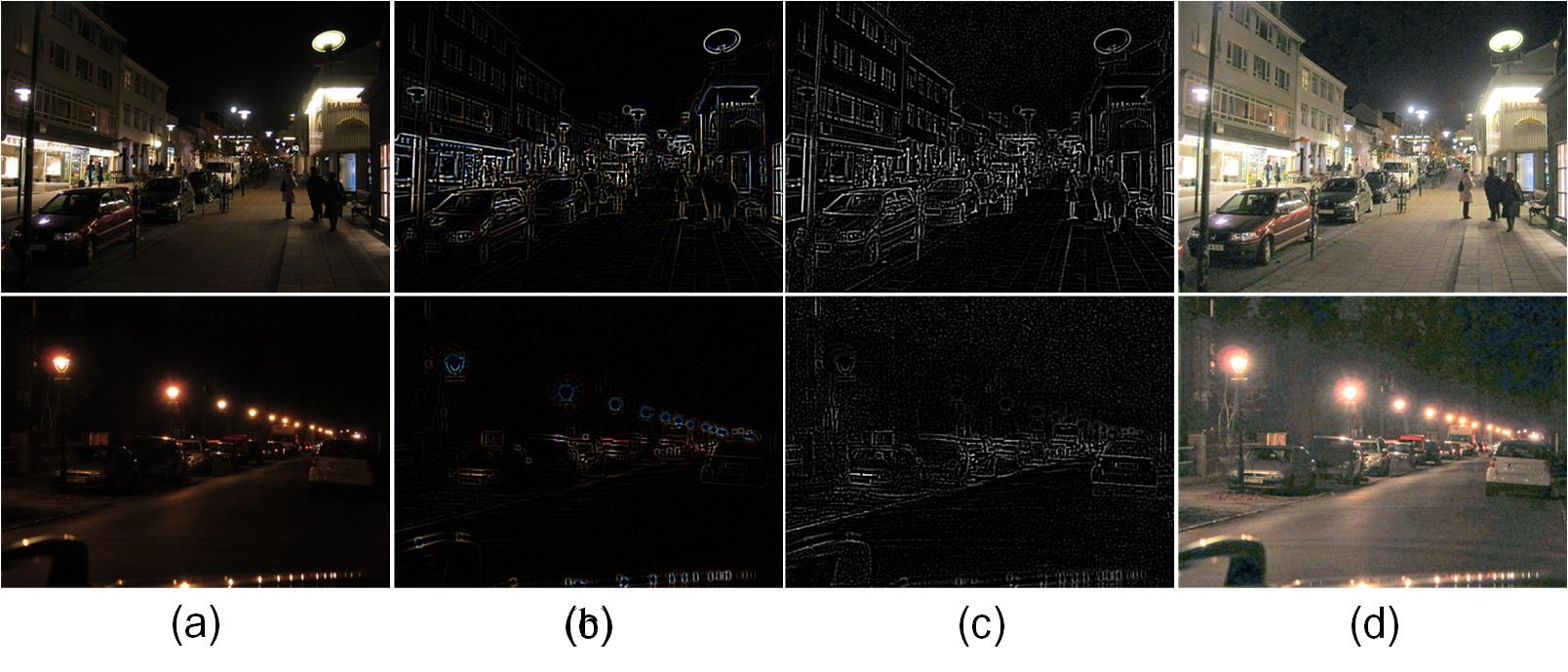}
        \caption{The examples of the enhanced results on gradient domain, from left to right: (a) low-light images, (b) LoG-based gradient feature map, (c) GEM-enhanced gradient map, and (d) final enhanced images.}
        \label{Figure_lap_coarse}
    \end{figure}

\subsection{Learning Methods}
    In recent years, deep learning \cite{lecun2015deep} has achieved widespread success in diverse fields of computer vision tasks, such as object detection, scene segmentation, and low-light image enhancement.  Based on the Retinex theory, many methods employed the CNN to formulate the decomposition and enhancement of low-light images, e.g., KinD \cite{zhang2019kindling}, RetinexNet \cite{wei2018deep}, RUAS \cite{liu2021retinex}, Uretinex-net \cite{wu2022uretinex} and LR3M \cite{9056796}. Meanwhile, many multi-branch networks \cite{lv2018mbllen, wang2020lightening, zhang2021learning, wang2022low,lu2022mtrbnet, guo2023low} were designed to tackle different subtasks in low lightness enhancement, e.g., noise reduction and color restoration. In addition to the supervised training, EnlightenGAN \cite{jiang2021enlightengan} and DRBN \cite{yang2021band} enlightened the darkness with semi-supervised network. LLFormer \cite{wang2023ultra} used vision transformer to achieve UHD low-light image enhancement. Although with considerable efforts, the running time of most previous works is not suitable for real-time UHD transportation surveillance. Besides, in transportation scenes, edge feature restoration is typically important, which was rarely considered. Lu \textit{et al}. \cite{lu2022low} proposed a gradient prior-aided neural network employing Laplacian and Sobel filters to guide the enhancement. However, these filters are sensitive to noise interference, which is harmful to image quality enhancement. In this paper, we employ the robust LoG operator to extract the gradient information and enhance it in the network to obtain better edge features.

\subsection{Applications in Transportation System}
    The efficient low-light image enhancement methods are necessary for nocturnal surveillance in ITS. Therefore, many efforts have been devoted to overcoming the restriction of poor illumination. For instance, a CycleGAN-based image enhancement method is proposed for railway inspections \cite{zheng2021image}, and an attention-guided lightweight generative adversarial network is designed for maritime video surveillance \cite{liu2022attention}. Guo \textit{et al}. \cite{guo2022lightweight} enlightened the darkness in maritime transportation scenes with a lightweight neural network. Besides, \cite{liu2021enhanced} and \cite{wu2022edge} have demonstrated the benefits of low-light enhancement for promoting the accuracy of higher-level image analysis tasks in ITS.

\section{Double Domain Guided Low-Light Image Enhancement Network}\label{Proposed Method}
    In this section, we first introduce the Laplacian of Gaussian Operator (LoG) in  Section \ref{pm-LoG}. The architecture of DDNet and the implementation details of the self-calibrated convolutions are then presented in Section \ref{pm-DDNet} and \ref{pm-attention module}. The joint loss function is introduced in Section \ref{pm-loss}.
    \begin{figure*}[t]
    	\centering
    	\setlength{\abovecaptionskip}{0.1cm}
    	\includegraphics[width=1.0\linewidth]{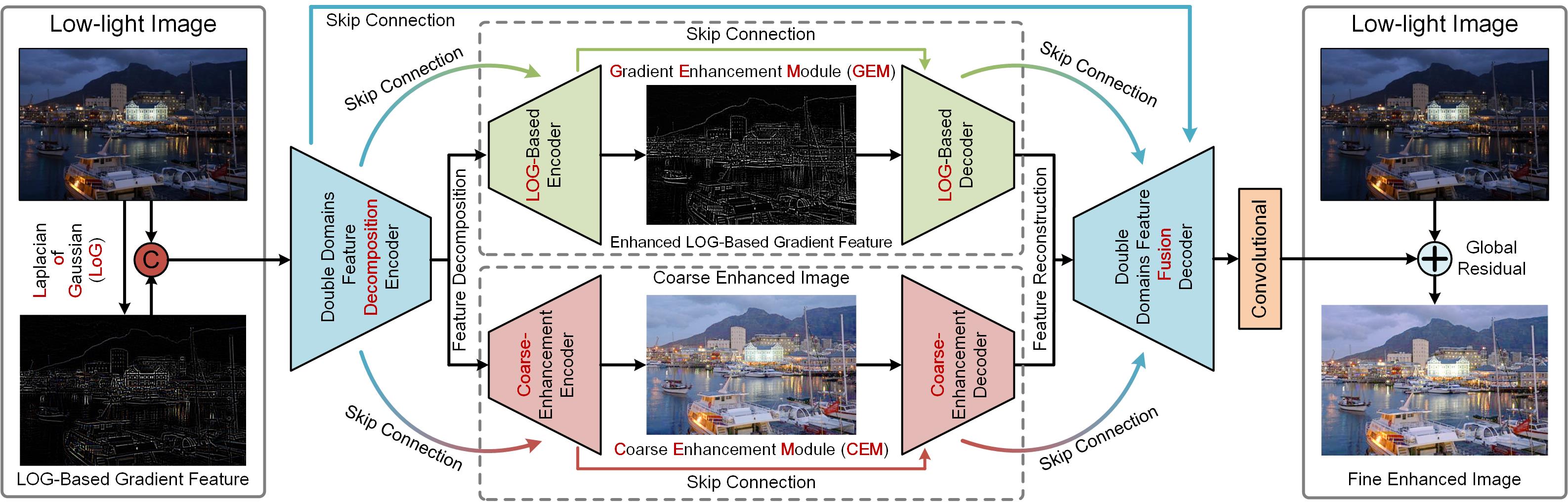}
    	\caption{The flowchart of our double domain guided low-light image enhancement network. The coarse enhancement module (CEM) and LoG-based gradient enhancement module (GEM) are embedded in the encoder-decoder structure to improve the image quality on separate domains. Moreover, the outputs of diversified decoders are constrained by the proposed joint loss function respectively.}
    	\label{flowchart}
    \end{figure*}

\subsection{Laplacian of Gaussian Operator}\label{pm-LoG}
    The transportation surveillance under low-light environments suffers from low brightness along with vague edge features, which causes knotty troubles to higher-level visual tasks in ITS \cite{chen2021high}. Therefore, it is necessary to take the restoration of edge features into consideration \cite{han2021using}. The Laplace operator is the sum of the second-order partial derivatives of the gray image function in the horizontal and vertical directions \cite{wang2007laplacian}. It responds to areas where the intensity changes rapidly and can be used to extract the image edge features. The Laplacian operator $L(u,v)$ corresponding to the intensity value $I$ of the image pixel can be given as follows
    \begin{equation}
        L(u, v)=\frac{\partial^{2} I}{\partial u^{2}}+\frac{\partial^{2} I}{\partial v^{2}}.
    \end{equation}

    A single image can be represented by a discrete set of pixel values. The gradient feature map can thus be generated through a second-order derivative discrete convolutional kernel $K_L$, which approximates the Laplacian operator, i.e.,
	\begin{equation}
    	K_L =\left[\begin{array}{ccc}
    	0 & +1 & 0 \\
    	+1 & -4 & +1 \\
    	0 & +1 & 0
    	\end{array}\right].
	\end{equation}
    However, the images captured in low-light environments commonly contain unwanted noise. The sensitivity to noise makes it challenging to accurately extract gradient features from low-light images. To this end, we first reduce the interference of noise on the image by Gaussian smoothing filtering, which can be expressed as follows
    \begin{equation}
        G_{\sigma}(u, v)=\frac{1}{2 \pi \sigma^{2}} \exp \left(-\frac{u^{2}+v^{2}}{2 \sigma^{2}}\right),
    \end{equation}
    where $\sigma$ is the Gaussian standard deviation. Benefiting from the associative property of the convolutional operation, we obtain a hybrid filter by convolving the Gaussian smoothing filter and Laplacian filter to generate LoG-based gradient features. The 2-D LoG function centered on zero with Gaussian standard deviation $\sigma$ is given by
    \begin{equation}
        LoG(u,v)=-\frac{1}{\pi \sigma^{4}}\left[1-\frac{u^{2}+v^{2}}{2 \sigma^{2}}\right] e^{-\frac{u^{2}+v^{2}}{2 \sigma^{2}}}.
    \end{equation}
    \begin{figure}[t]
        	\centering
        	\setlength{\abovecaptionskip}{0.1cm}
        	\includegraphics[width=1.00\linewidth]{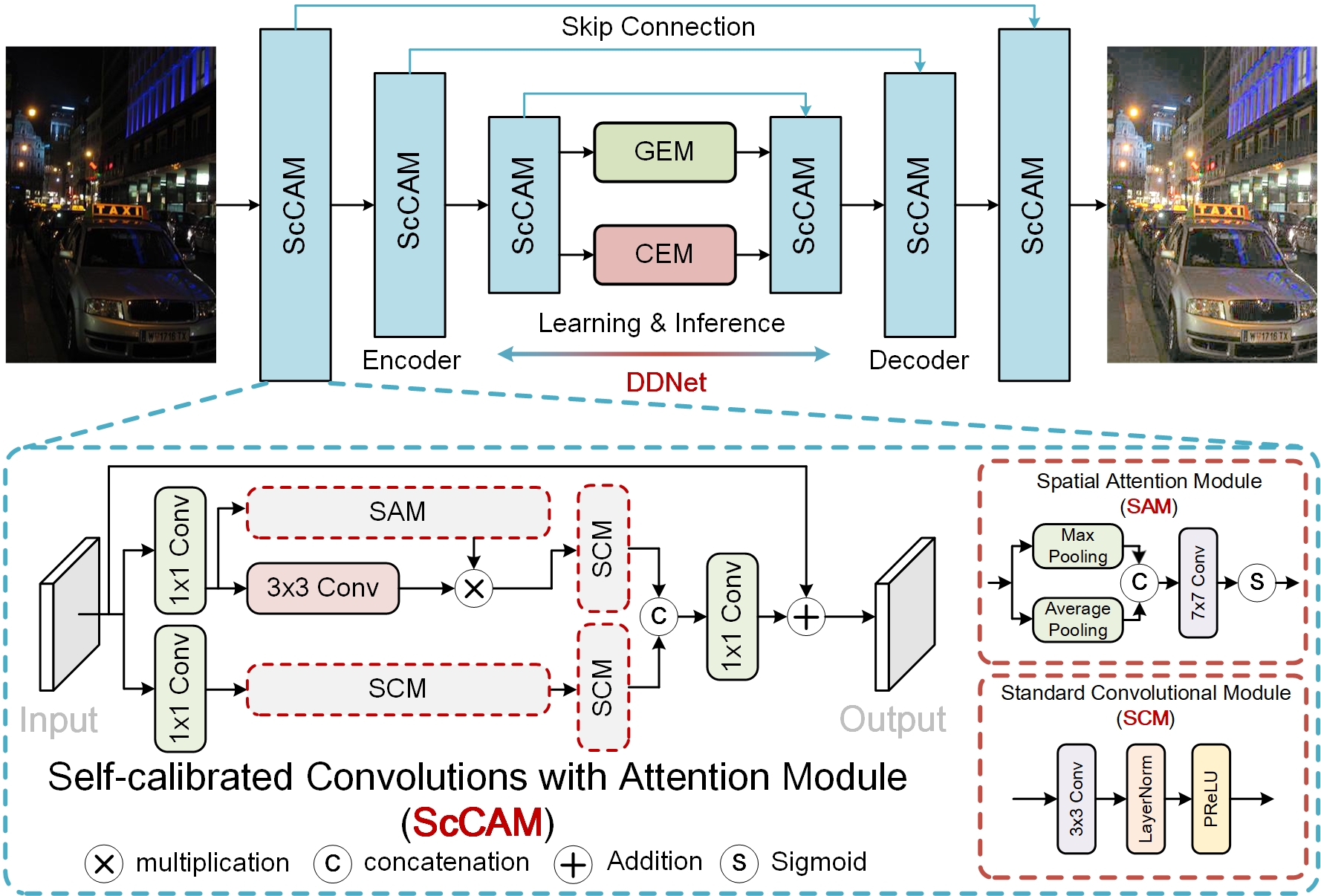}
        	\caption{The sketch map of the encoder-decoder structure, which employs the Self-calibrated Convolutions with Attention Module (ScCAM) for spatial and attention feature extraction.}
        	\label{Figure03_ScCAM}
    \end{figure}
    The convolutional kernel of LoG is small, and the kernel parameters are pre-calculated, which brings little computational burden. In this work, the convolutional kernel parameters of LoG can be given as follows
	\begin{equation}
    	K_{LoG} =\left[\begin{array}{ccccc}
    	0 & 0 & +1 & 0 & 0 \\
    	0 & +1 & +2 & +1 & 0 \\
    	+1 & +2 & -16 & +2 & +1 \\
    	0 & +1 & +2 & +1 & 0 \\
    	0 & 0 & +1 & 0 & 0 \\
    	\end{array}\right].
	\end{equation}
    In the network, we first generate the gradient map of the low-light image via the LoG-based operator, which will be then enhanced in the GEM, as shown in Fig. \ref{Figure_lap_coarse}.

\subsection{Network Architecture}\label{pm-DDNet}
    An ordinary neural network can not simultaneously and accurately generate the normal-light image and gradient feature map from the low-light image. We thus use multi-stage architecture to perform fusion-decomposition-fusion on the color and gradient domains. For the sake of better understanding, Fig. \ref{flowchart} depicts the architecture of our DDNet. Specifically, we first concatenate low-light images and their corresponding LoG-based gradient feature maps and feed them into the network. The proposed architecture includes six self-calibrated convolutions with attention modules (ScCAM) in the peripheral en-decoder, GEM and CEM, respectively. As introduced in Section. \ref{pm-attention module}, the ScCAMs leverage spatial attention to identify valuable information locations within the feature maps, which are then utilized for self-calibration convolutions. This enables the convolutional modules to extract more important features without incurring additional computational costs. Additionally, the feature maps share similar structures, e.g., the same size (width and height) and intensity range ($[0, 255]$), allowing ScCAM to effectively extract and enhance the spatial features on gradient and color domains simultaneously. Therefore, the potential spatial features of gradient and color domains are effectively mined and enhanced during the en-decoding in GEM and CEM. The outputs of GEM and CEM, as well as the outputs of previous encoders are then fed to the final feature fusion decoder, which reconstructs the normal-light image based on the fused feature map. It is noted that during the training process, the enhanced gradient and color maps are generated by their respective decoders, which are constrained by individual loss functions to guarantee the restoration of both gradient and color information, as introduced in Section \ref{pm-loss}. Due to the comprehensive enhancement on double domains with GEM and CEM, the proposed DDNet restores the low-light image with clear edges and natural colors. 

\subsection{ScCAM}\label{pm-attention module}
    To reduce the computational parameters, the majority of deep learning-based lightweight low-light enhancement networks extract hierarchical features progressively. However, this strategy leads to the insufficient utilization of low-frequency information, which results in poor performance on image detail restoration. Meanwhile, the self-calibrated convolutions (SCCs) perform satisfactorily in a variety of low-level and higher-level vision tasks \cite{zou2022self}. SCCs can efficiently extract multi-domain and multi-scale feature information to guide the enhancement processing without additional computational effort. In this section, we propose the ScCAM to conduct the encoder-decoder structures. It mainly consists of two parts (i.e., the upper and lower branches), as shown in Fig. \ref{Figure03_ScCAM}. In particular, the upper part computes the attention information by introducing the spatial attention module, which can be expressed as follows
    \begin{equation}
        y_{\text {upper}}=F_{\operatorname{scm}}\left(M\left(F_{\text {sam }}\left(f^{1 \times 1}\left(x_{\text {in}}\right)\right) ; f^{3 \times 3}\left(f^{1 \times 1}\left(x_{\text{in}}\right)\right)\right)\right),
    \end{equation}
    where $x_{in}$, $f^{1 \times 1}$, $f^{3 \times 3}$, $F_{sam}$, $M( \cdot ; \cdot )$, and $F_{scm}$ represent the input of convolutional layer, the convolutional operation with 1×1 kernel size, the convolutional operation with 3×3 kernel size, the spatial attention module, the multiplication function, and the standard convolution module, respectively. In addition, the lower part uses the standard convolution module to recover the spatial domain information, which can be expressed as follows
    \begin{equation}
        y_{\text {lower }}=F_{\operatorname{scm}}\left(F_{\operatorname{scm}}\left(f^{1 \times 1}\left(x_{\text {in }}\right)\right)\right).
    \end{equation}

    The output features of these two parts are then concatenated together and fed into a 1×1 convolution layer for information fusion. To speed up model training, the local residual path is employed to generate the final output feature. The output ($y_{ScCAM}$) of ScCAM can be thus yielded by
    \begin{equation}
        y_{ScCAM}=f^{1 \times 1}\left(y_{upper} ; y_{lower}\right)+x_{i n},
    \end{equation}
    where $( \cdot ; \cdot )$ represents the concatenation operation. 

\subsubsection{Spatial Attention Module}
    In the process of low-light image enhancement, the complexity of scene information increases the difficulty of enhancement. Considering the human visual cerebral cortex, applying the attention mechanism can analyze complex scene information more quickly and effectively. The spatial attention module is beneficial for analyzing where the valuable information on the feature map is, which contributes to focusing more precisely on the feature map's valuable information. As shown in Fig. \ref{Figure03_ScCAM}, to achieve spatial attention, we first use the average pooling and max pooling in the channel dimension. The feature maps are then concatenated and fed into a convolution layer with 7 $\times$ 7 kernel to generate the final spatial attention feature map. The spatial attention function can be expressed as follows
    \begin{equation}
        \begin{aligned}
        F_{sam}(\mathcal{I}) 
        &=S\left(f^{7 \times 7}\left(F_{a v g}^s(\mathcal{I}) ; F_{\max }^s(\mathcal{I})\right)\right),
        \end{aligned}
    \end{equation} 
    where $\mathcal{I}$, $F_{a v g}^s$, $F_{m a x}^s$, $f^{7 \times 7}$, and $S(\cdot)$ represent the inputs of spatial attention module, average pooling, max pooling, the convolutional operation with 7×7 kernel size, and the sigmoid function, respectively.

\subsubsection{Standard Convolution Module}
	In the standard convolution module, the convolution layer is first employed to guarantee the learning ability. Layer normalization (LN) is independent of batch size, which reduces the computational complexity when calculating normalization statistics. Furthermore, the Parametric Rectified Linear Unit (PReLU) is employed to perform nonlinear activation on the normalized data, which improves the generalization ability of the network in complex low-light scenes. The standard convolution function can be generated as follows
	\begin{equation}
		F_{scm}(w)=PR(LN(f^{3 \times 3}(w))),
	\end{equation}
    where $w$, $LN(\cdot)$, and $PR(\cdot)$ represent the inputs of the standard convolution module, layer normalization, and parametric rectified linear unit, respectively.
    \setlength{\tabcolsep}{1pt}
    \begin{table}[t]
        \scriptsize
        \centering
        \caption{The quantitative comparison between our method and state-of-the-arts on the LOL test dataset \cite{wei2018deep}. The best three results are highlighted in {\color{red}red}, {\color{blue}blue}, and {\color{green}green} colors. $\uparrow$ and $\downarrow$ represent that higher or lower values indicate better results, respectively.}
        \begin{tabular}{l|ccccc}
        \hline
        Method                                                    & PSNR $\uparrow$                               & SSIM $\uparrow$                                & NIQE $\downarrow$                            & PIQE $\downarrow$                            & LPIPS $\downarrow$ \\ \hline \hline
        HE \cite{HE}                              & 14.80$\pm$3.09                               & 0.386$\pm$0.085                               & 8.48$\pm$1.17                               & 15.65$\pm$2.98                              &  0.340$\pm$0.146          \\
        NPE \cite{improving1}                     & 16.97$\pm$2.87                               & 0.474$\pm$0.116                               & 8.44$\pm$1.21                               & 14.88$\pm$3.74                              &  0.339$\pm$0.166          \\
        LIME \cite{improving3}                   & 17.18$\pm$3.89                               & 0.556$\pm$0.113                               & 4.99$\pm$1.00                               & 22.53$\pm$7.70                              & 0.343$\pm$0.171          \\
        JIEP \cite{cai2017joint}                  & 12.05$\pm$3.92                               & 0.510$\pm$0.134                               & 6.87$\pm$0.85                               & \color{blue}5.49$\pm$1.71 &  0.296$\pm$0.180          \\
        CRM \cite{ying2017new}                    & 17.20$\pm$4.87                               & 0.618$\pm$0.109                               & 7.69$\pm$1.04                               & 9.57$\pm$3.75                               & 0.175$\pm$0.087          \\
        Dong \cite{dong2010fast}                  & 16.72$\pm$3.99                               & 0.472$\pm$0.099                               & 8.32$\pm$1.10                            & 14.92$\pm$2.94                              & 0.217$\pm$0.010          \\
        BIMEF \cite{ying2017bio}                  & 13.88$\pm$4.45                               & 0.594$\pm$0.121                               & 7.52$\pm$1.09                               & 7.61$\pm$3.53                               & 0.174$\pm$0.089          \\ 
        DeHz \cite{jiang2013night}                  & 15.69$\pm$4.59                               & 0.54$\pm$0.123                               & 7.64$\pm$1.16                            & 10.86$\pm$3.32                              & 0.186$\pm$0.100          \\\hline
        RetinexNet \cite{wei2018deep}             & 17.12$\pm$2.58                               & 0.596$\pm$0.086                               & 6.40$\pm$0.93                               & 18.42$\pm$4.42                              &  0.204$\pm$0.069          \\
        MBLLEN \cite{lv2018mbllen}                & 17.86$\pm$3.55                               & 0.727$\pm$0.061                               & \color{green}4.36$\pm$0.67                               & 15.02$\pm$7.04                              &  0.131$\pm$0.032          \\
        KinD \cite{zhang2019kindling}             & 17.71$\pm$3.30                               & 0.777$\pm$0.101                               & 4.69$\pm$1.26                               & 16.11$\pm$9.59                              & \color{blue}0.096$\pm$0.054         \\
        EG \cite{jiang2021enlightengan} & 17.45$\pm$4.49                               & 0.644$\pm$0.111                               & 4.81$\pm$0.84                               & 9.42$\pm$3.38                               &  0.185$\pm$0.110          \\
        DLN \cite{wang2020lightening}             & 19.26$\pm$4.05                               & 0.695$\pm$0.089                               & 6.09$\pm$0.89                               & \color{green}7.50$\pm$3.01                               & 0.153$\pm$0.068          \\
        Zero \cite{guo2020zero}                   & 14.86$\pm$4.27                               & 0.559$\pm$0.125                               & 7.77$\pm$1.11                               & 10.34$\pm$3.73                              &  0.189$\pm$0.010         \\
        StableLLVE \cite{zhang2021learning}       & 17.36$\pm$3.41                               & 0.740$\pm$0.089                               & 4.37$\pm$0.61                               & 7.57$\pm$4.99                               & 0.175$\pm$0.068          \\
        RUAS \cite{liu2021retinex}                & 16.40$\pm$4.39                               & 0.500$\pm$0.114                               & 6.34$\pm$1.17                               & \color{red}3.26$\pm$1.07   & 0.149$\pm$0.078          \\
        LLFlow \cite{wang2022low}                 & \color{green}19.34$\pm$3.01                               & \color{red}0.840$\pm$0.066   & 5.46$\pm$1.09                               & 29.63$\pm$11.49                             & \color{red}0.077$\pm$0.036          \\
        MTRBNet \cite{lu2022mtrbnet}              & \color{blue}21.21$\pm$3.69  & \color{green}0.785$\pm$0.083 & \color{blue}3.97$\pm$0.31 & 15.73$\pm$7.81                              & 0.139$\pm$0.065          \\ 
        SCI \cite{ma2022toward}               & 14.82$\pm$4.41  & 0.639$\pm$0.169 & 7.88$\pm$1.10 & 10.25$\pm$2.41                              & 0.203$\pm$0.103          \\ \hline
        DDNet                                                      & \color{red}21.86$\pm$4.36   & \color{blue}0.832$\pm$0.078  & \color{red}3.38$\pm$0.54   & 9.62$\pm$6.18                               & \color{green}0.108$\pm$0.046          \\ \hline
        \end{tabular}\label{Table02_LOL}
    \end{table}
    %
    %
	% We verified the quantitative and visual performance of each unit through an ablation study in section \ref{Ablation Study}. %Experimental results show that only when the spatial attention module and the standard convolution module collaborate on low-light learning and mapping, the best enhancement performance is obtained.
	%
\subsection{Loss Function}\label{pm-loss}
	To effectively constrain each component of the DDNet, we propose a joint loss function $\mathcal{L}_{total}$ consisting of Laplacian-based gradient consistency loss $\mathcal{L}_{\text{Lap}}$, coarse enhancement loss $\mathcal{L}_{\text{Coarse}}$, and final enhancement loss $\mathcal{L}_{\text{Final}}$, which can be expressed as follows
	\begin{equation}\label{eq:l_total}
	    \mathcal{L}_{total} = \omega_{1}\mathcal{L}_{\text{Lap}} + \omega_{2}\mathcal{L}_{\text{Coarse}} +\omega_{3}\mathcal{L}_{\text{Final}},
	\end{equation}
	where $\omega_{1}$, $\omega_{2}$, and $\omega_{3}$ are the weights of each loss, which are set to 0.2, 0.2, and 0.6, respectively. The GEM and CEM are proposed to enhance the gradient and color features, respectively, which are constrained by the $\ell_2$ loss function. The $\mathcal{L}_{\text{Lap}}$ and $\mathcal{L}_{\text{Coarse}}$ can be given as follows
	\begin{equation}\label{eq:l_lap}
    	\mathcal{L}_{\text{Lap}} =\frac{1}{N}\sum\limits_{p=1}^N\sum\limits_{i=1}^1||\hat{I}_{i}^{l}(p)-I_{i}^{l}(p)||^2,
	\end{equation}
	\begin{equation}\label{eq:l_coarse}
    	\mathcal{L}_{\text{Coarse}} =\frac{1}{N}\sum\limits_{p=1}^N\sum\limits_{i=1}^3||\hat{I}_{i}^{c}(p)-I_{i}^{c}(p)||^2,
	\end{equation}
	where $N$ is the number of pixels, $\hat{I}_i^l(p)$ and $I_i^l(p)$ are the $i$-th color channel of pixel $p$ in the gradient map of low-light image and ground truth, respectively. $\hat{I}_i^c(p)$ and $I_i^{c}(p)$ represent the corresponding values on the color domain.
	To finely fuse the gradient and coarse enhancement features, we use the structural similarity (SSIM) \cite{wang2004image} as the constraint of the final enhancement to further refine the learning and mapping, i.e.,
	\begin{equation}\label{eq:l_fine}
    	\mathcal{L}_{\text{Final}} = 1-\sum\limits_{i=1}^3ssim(\hat{I}_i^{f},I_i),
	\end{equation}
    where $\hat{I}_i^{f}$ is the final fine enhancement image, and $I_i$ is the ground truth. $ssim(\cdot, \cdot)$ calculates the structural similarity consisting of the aspects of color, structure, and contrast.

\section{Experiments and Analysis}\label{Experiments}
    In this section, the experimental details are first introduced, which include datasets, evaluation metrics, and running platform. To clearly demonstrate the superiority of DDNet, qualitative and quantitative comparisons with several state-of-the-art methods on standard and transportation-related datasets are then presented. To validate the rationality of the network, we conduct ablation experiments on each module. The experiments on running time, object detection, and scene segmentation are finally conducted, which demonstrate practical contributions of the proposed method to real-time UHD transportation surveillance in ITS.

\subsection{Implementation Details}
\subsubsection{Datasets}
    It is commonly intractable to capture the real-world low/normal-light image pairs, which brings great challenges for data-driven image enhancement networks. Therefore, to improve the robustness of our DDNet to the complex natural environments, we utilize the real-captured and synthesized low-light images simultaneously. The most commonly used dataset is LOL \cite{wei2018deep}, which contains 1500 pairs of low-light images. Among them, 500 pairs are captured in real scenes, and the rest are synthesized with the adaption of the Y channel in YCbCr image through the interface from Adobe Light-room software \footnote{The hyperparameters of Adobe Light-room software: Exposure ($-5 + 5F$), Highlights ($50 \min{\{Y, 0.5\}} + 75$), Shadows ($-100\min{\{Z, 0.5\}}$), Vibrance ($-75 + 75F$), and Whites ($16(5-5F)$). It is noted that the $X$, $Y$, and $Z$ are the variable obeys uniform random distribution $\mathcal{U}(0, 1)$, and $F = X^2$.}.
    
    Besides LOL, to improve the enhancement effect on transportation surveillance scenes, we select 1000 clear outdoor images from the PASCAL VOC 2007 \cite{everingham2010pascal}, COCO \cite{lin2014microsoft}, as well as DETRAC \cite{wen2020ua} datasets and synthesize the low-light images with another method, which multiplies a specific coefficient to all image pixels. The synthesized image $L(x)$ can be generated by
        \begin{equation}
            L(x)=C(x)m(x),
            \label{synthetic}
        \end{equation}
    where $C(x)$ is the clear image, and $m(x)$ is the coefficient, which is a random number between 0.1 and 0.9. To prove the generalization ability of DDNet, besides evaluation on the LOL dataset, we also select representative low-light images from DICM \cite{guo2020zero}, LIME \cite{improving3}, MEF \cite{lee2011power}, and TMDIED dataset for testing.
    \setlength{\tabcolsep}{3.0pt}
    \begin{table}[t]
        \scriptsize
        \centering
        \caption{The quantitative comparison of NIQE between our method and the state-of-the-arts on DICM \cite{guo2020zero}, LIME \cite{improving3}, MEF \cite{lee2011power}, and TMDIED dataset. The best three results are highlighted in {\color{red}red}, {\color{blue}blue}, and {\color{green}green} colors.}
        \label{table_niqe}
        \begin{tabular}{l|clclclclc}
        \hline
         Method          & \multicolumn{2}{c}{DICM}              & \multicolumn{2}{c}{LIME}              & \multicolumn{2}{c}{MEF}               & \multicolumn{2}{c}{TMDIED}            & Average                                  \\ \hline \hline
        
        HE \cite{HE}         & \multicolumn{2}{c}{3.68$\pm$1.33}        & \multicolumn{2}{c}{4.44$\pm$3.14}        & \multicolumn{2}{c}{3.64$\pm$1.30}        & \multicolumn{2}{c}{4.83$\pm$1.37}        & 4.58$\pm$1.51                               \\
        NPE \cite{improving1}        & \multicolumn{2}{c}{3.68$\pm$1.26}        & \multicolumn{2}{c}{3.93$\pm$1.93}        & \multicolumn{2}{c}{3.54$\pm$1.15}        & \multicolumn{2}{c}{4.67$\pm$1.24}        & 4.39$\pm$1.37                               \\
                          LIME \cite{improving3}        & \multicolumn{2}{c}{3.63$\pm$0.82}        & \multicolumn{2}{c}{4.71$\pm$2.05}        & \multicolumn{2}{c}{4.25$\pm$1.10}        & \multicolumn{2}{c}{3.81$\pm$0.86}        & 3.82$\pm$0.95                               \\
        JIEP \cite{cai2017joint}       & \multicolumn{2}{c}{3.77$\pm$0.85}        & \multicolumn{2}{c}{\color{green}3.74$\pm$1.60}   & \multicolumn{2}{c}{3.39$\pm$1.09}        & \multicolumn{2}{c}{4.31$\pm$1.12}        & 4.06$\pm$1.18                               \\
        CRM \cite{ying2017new}        & \multicolumn{2}{c}{3.77$\pm$1.04}        & \multicolumn{2}{c}{3.87$\pm$1.72}        & \multicolumn{2}{c}{\color{blue}3.27$\pm$1.13}    & \multicolumn{2}{c}{4.58$\pm$1.22}        & 4.29$\pm$1.31                               \\
        Dong \cite{dong2010fast}       & \multicolumn{2}{c}{3.56$\pm$1.18}        & \multicolumn{2}{c}{4.11$\pm$1.99}        & \multicolumn{2}{c}{4.11$\pm$1.13}        & \multicolumn{2}{c}{5.15$\pm$1.15}        & 4.89$\pm$1.27                               \\
        BIMEF \cite{ying2017bio}      & \multicolumn{2}{c}{\color{green}3.54$\pm$1.00}   & \multicolumn{2}{c}{3.86$\pm$1.63}        & \multicolumn{2}{c}{3.33$\pm$1.14}        & \multicolumn{2}{c}{4.47$\pm$1.21}        & 4.22$\pm$1.28                               \\
        DeHz \cite{jiang2013night}       & \multicolumn{2}{c}{3.61$\pm$0.98}        & \multicolumn{2}{c}{3.89$\pm$1.79}        & \multicolumn{2}{c}{3.48$\pm$1.41}        & \multicolumn{2}{c}{4.29$\pm$1.16}        & 4.07$\pm$1.24                               \\ \hline
        RetinexNet \cite{wei2018deep} & \multicolumn{2}{c}{3.58$\pm$1.21}        & \multicolumn{2}{c}{5.33$\pm$4.06}        & \multicolumn{2}{c}{4.56$\pm$1.25}        & \multicolumn{2}{c}{4.13$\pm$0.89}        & 4.24$\pm$1.21                               \\
        MBLLEN \cite{lv2018mbllen}     & \multicolumn{2}{c}{3.56$\pm$0.88}        & \multicolumn{2}{c}{4.49$\pm$1.69}        & \multicolumn{2}{c}{4.74$\pm$2.32}        & \multicolumn{2}{c}{4.07$\pm$1.02}        & 4.09$\pm$1.15                               \\
        KinD \cite{zhang2019kindling}      & \multicolumn{2}{c}{3.73$\pm$1.15}        & \multicolumn{2}{c}{4.71$\pm$3.27}        & \multicolumn{2}{c}{3.82$\pm$1.04}        & \multicolumn{2}{c}{\color{green}3.39$\pm$0.84}   & \color{green}3.48$\pm$1.10 \\
        EnlightenGAN \cite{jiang2021enlightengan}         & \multicolumn{2}{c}{3.80$\pm$0.54}        & \multicolumn{2}{c}{\color{red}3.65$\pm$1.54}     & \multicolumn{2}{c}{\color{red}3.05$\pm$0.74}     & \multicolumn{2}{c}{3.88$\pm$0.94}        & 3.68$\pm$0.98                               \\
        DLN \cite{wang2020lightening}        & \multicolumn{2}{c}{3.77$\pm$0.74}        & \multicolumn{2}{c}{3.88$\pm$1.74}        & \multicolumn{2}{c}{3.36$\pm$1.13}        & \multicolumn{2}{c}{4.54$\pm$1.23}        & 4.24$\pm$1.30                               \\
        Zero \cite{guo2020zero}       & \multicolumn{2}{c}{3.75$\pm$1.00}        & \multicolumn{2}{c}{3.78$\pm$1.86}        & \multicolumn{2}{c}{\color{green}3.28$\pm$1.03}   & \multicolumn{2}{c}{4.62$\pm$1.20}        & 4.32$\pm$1.31                               \\
        StableLLVE \cite{zhang2021learning} & \multicolumn{2}{c}{3.77$\pm$0.68}        & \multicolumn{2}{c}{4.23$\pm$1.55}        & \multicolumn{2}{c}{3.88$\pm$1.05}        & \multicolumn{2}{c}{\color{blue}3.23$\pm$0.71}    & \color{blue}3.34$\pm$0.80  \\
        RUAS \cite{liu2021retinex}       & \multicolumn{2}{c}{3.78$\pm$1.50}        & \multicolumn{2}{c}{5.41$\pm$2.00}        & \multicolumn{2}{c}{5.42$\pm$1.12}        & \multicolumn{2}{c}{5.92$\pm$1.84}        & 5.80$\pm$1.77                               \\
        LLFlow \cite{wang2022low}     & \multicolumn{2}{c}{3.79$\pm$0.75}        & \multicolumn{2}{c}{4.47$\pm$1.88}        & \multicolumn{2}{c}{3.96$\pm$1.00}        & \multicolumn{2}{c}{3.70$\pm$0.96}        & 3.69$\pm$0.99                               \\
        MTRBNet \cite{lu2022mtrbnet}    & \multicolumn{2}{c}{\color{blue}3.31$\pm$0.74}    & \multicolumn{2}{c}{4.32$\pm$1.04}        & \multicolumn{2}{c}{4.50$\pm$0.93}        & \multicolumn{2}{c}{3.91$\pm$0.73}        & 3.98$\pm$0.77                               \\ 
        SCI \cite{ma2022toward}    & \multicolumn{2}{c}{4.09$\pm$1.35}    & \multicolumn{2}{c}{{\color{blue}3.66$\pm$1.02}}        & \multicolumn{2}{c}{3.63$\pm$1.05}        & \multicolumn{2}{c}{5.13$\pm$1.42}        & 4.80$\pm$1.50                               \\ \hline
        DDNet      & \multicolumn{2}{c}{\color{red}2.98$\pm$0.68}     & \multicolumn{2}{c}{3.76$\pm$1.53}    & \multicolumn{2}{c}{3.29$\pm$0.90}        & \multicolumn{2}{c}{\color{red}3.09$\pm$0.69}     & \color{red}3.11$\pm$0.75   \\ \hline
        \end{tabular}
    \end{table}
    \setlength{\tabcolsep}{0.5pt}
    \begin{table}[t]
        \scriptsize
        \centering
        \caption{The quantitative comparison of PIQE between our method and the state-of-the-arts on DICM \cite{guo2020zero}, LIME \cite{improving3}, MEF \cite{lee2011power}, and TMDIED dataset. The best three results are highlighted in {\color{red}red}, {\color{blue}blue}, and {\color{green}green} colors.}
        \label{table_piqe}
        \begin{tabular}{l|clclclclc}
        \hline
        Method    & \multicolumn{2}{c}{DICM}            & \multicolumn{2}{c}{LIME}              & \multicolumn{2}{c}{MEF}             & \multicolumn{2}{c}{TMDIED}          & Average                                  \\ \hline \hline
        HE \cite{HE}         & \multicolumn{2}{c}{10.80$\pm$5.52}     & \multicolumn{2}{c}{12.88$\pm$7.87}       & \multicolumn{2}{c}{9.78$\pm$1.84}      & \multicolumn{2}{c}{10.04$\pm$3.89}     & 10.24$\pm$4.29                              \\
        NPE \cite{improving1}        & \multicolumn{2}{c}{9.15$\pm$4.97}      & \multicolumn{2}{c}{11.46$\pm$8.97}       & \multicolumn{2}{c}{8.68$\pm$3.48}      & \multicolumn{2}{c}{8.65$\pm$3.59}      & 8.82$\pm$4.10                               \\
        LIME \cite{improving3}       & \multicolumn{2}{c}{16.98$\pm$7.41}     & \multicolumn{2}{c}{19.35$\pm$12.65}      & \multicolumn{2}{c}{25.73$\pm$11.87}    & \multicolumn{2}{c}{12.00$\pm$6.37}     & 7.68$\pm$4.37                               \\
        JIEP \cite{cai2017joint}       & \multicolumn{2}{c}{\color{blue}6.30$\pm$3.04}  & \multicolumn{2}{c}{11.42$\pm$9.68}       & \multicolumn{2}{c}{9.78$\pm$8.28}      & \multicolumn{2}{c}{\color{green}6.63$\pm$3.09} & 13.75$\pm$8.00                              \\
        CRM \cite{ying2017new}        & \multicolumn{2}{c}{8.87$\pm$4.26}      & \multicolumn{2}{c}{11.91$\pm$10.30}      & \multicolumn{2}{c}{\color{green}8.12$\pm$4.13}      & \multicolumn{2}{c}{8.13$\pm$3.51}      & \color{blue}6.92$\pm$4.04  \\
        Dong \cite{dong2010fast}       & \multicolumn{2}{c}{10.10$\pm$4.29}     & \multicolumn{2}{c}{11.38$\pm$6.58}       & \multicolumn{2}{c}{9.07$\pm$3.09}      & \multicolumn{2}{c}{10.46$\pm$3.36}     & 8.37$\pm$4.09                               \\
        BIMEF \cite{ying2017bio}      & \multicolumn{2}{c}{7.94$\pm$4.14}      & \multicolumn{2}{c}{11.70$\pm$11.04}      & \multicolumn{2}{c}{8.33$\pm$6.20}      & \multicolumn{2}{c}{6.82$\pm$3.45}      & 10.36$\pm$6.63                              \\
        DeHz \cite{jiang2013night}       & \multicolumn{2}{c}{8.47$\pm$3.68}      & \multicolumn{2}{c}{11.51$\pm$9.70}       & \multicolumn{2}{c}{8.53$\pm$7.17}      & \multicolumn{2}{c}{8.12$\pm$3.42}      & 7.24$\pm$4.28                               \\ \hline
        RetinexNet \cite{wei2018deep} & \multicolumn{2}{c}{13.61$\pm$4.76}     & \multicolumn{2}{c}{13.91$\pm$8.33}       & \multicolumn{2}{c}{12.35$\pm$5.13}     & \multicolumn{2}{c}{14.08$\pm$4.48}     & 13.91$\pm$4.72                              \\
        MBLLEN \cite{lv2018mbllen}     & \multicolumn{2}{c}{17.78$\pm$8.99}     & \multicolumn{2}{c}{22.05$\pm$16.88}      & \multicolumn{2}{c}{27.28$\pm$13.75}    & \multicolumn{2}{c}{14.01$\pm$8.19}     & 15.58$\pm$9.64                              \\
        KinD \cite{zhang2019kindling}       & \multicolumn{2}{c}{15.74$\pm$7.97}     & \multicolumn{2}{c}{17.00$\pm$10.96}      & \multicolumn{2}{c}{19.40$\pm$11.77}    & \multicolumn{2}{c}{9.00$\pm$4.90}      & 10.86$\pm$7.07                              \\
        EnlightenGAN \cite{jiang2021enlightengan}         & \multicolumn{2}{c}{\color{red}6.23$\pm$2.21}   & \multicolumn{2}{c}{\color{red}9.51$\pm$8.58}    & \multicolumn{2}{c}{\color{blue}7.95$\pm$5.98} & \multicolumn{2}{c}{7.28$\pm$2.98}      & \color{green}7.23$\pm$3.46 \\
        DLN \cite{wang2020lightening}        & \multicolumn{2}{c}{7.35$\pm$2.88} & \multicolumn{2}{c}{\color{blue}10.71$\pm$10.81} & \multicolumn{2}{c}{\color{red}7.51$\pm$4.34}  & \multicolumn{2}{c}{7.28$\pm$3.46}      & 7.42$\pm$3.91                               \\
        Zero \cite{guo2020zero}       & \multicolumn{2}{c}{8.74$\pm$3.97}      & \multicolumn{2}{c}{\color{green}11.17$\pm$8.51}       & \multicolumn{2}{c}{8.17$\pm$3.60}      & \multicolumn{2}{c}{8.40$\pm$3.49}      & 8.53$\pm$3.84                               \\
        StableLLVE \cite{zhang2021learning} & \multicolumn{2}{c}{10.28$\pm$5.08}     & \multicolumn{2}{c}{13.72$\pm$13.06}      & \multicolumn{2}{c}{19.75$\pm$11.89}    & \multicolumn{2}{c}{\color{red}5.82$\pm$3.87}  & 7.52$\pm$6.38                               \\
        RUAS \cite{liu2021retinex}      & \multicolumn{2}{c}{14.67$\pm$7.20}     & \multicolumn{2}{c}{17.31$\pm$13.52}      & \multicolumn{2}{c}{17.25$\pm$9.28}     & \multicolumn{2}{c}{9.51$\pm$6.67}      & 10.98$\pm$7.67                              \\
        LLFlow \cite{wang2022low}     & \multicolumn{2}{c}{9.78$\pm$4.63}      & \multicolumn{2}{c}{18.77$\pm$13.21}      & \multicolumn{2}{c}{18.90$\pm$12.19}    & \multicolumn{2}{c}{17.45$\pm$6.47}     & 10.75$\pm$7.38                              \\
        MTRBNet \cite{lu2022mtrbnet}    & \multicolumn{2}{c}{25.99$\pm$11.02}    & \multicolumn{2}{c}{23.19$\pm$16.31}      & \multicolumn{2}{c}{27.76$\pm$9.11}     & \multicolumn{2}{c}{18.04$\pm$9.73}     & 19.95$\pm$10.69                             \\ 
        SCI \cite{ma2022toward}     & \multicolumn{2}{c}{16.94$\pm$2.54}    & \multicolumn{2}{c}{12.11$\pm$10.05}      & \multicolumn{2}{c}{8.85$\pm$3.99}     & \multicolumn{2}{c}{10.66$\pm$4.31}     & 12.05$\pm$7.58                             \\ \hline
        DDNet      & \multicolumn{2}{c}{\color{green}6.48$\pm$2.59}       & \multicolumn{2}{c}{13.47$\pm$12.58}    & \multicolumn{2}{c}{9.55$\pm$7.87}   & \multicolumn{2}{c}{\color{blue}6.21$\pm$3.24}    & \color{red}6.71$\pm$4.41    \\ \hline
        \end{tabular}
    \end{table}
    \begin{figure*}[t]
    	\centering
    	\setlength{\abovecaptionskip}{0.1cm}
    	\includegraphics[width=1.0\linewidth]{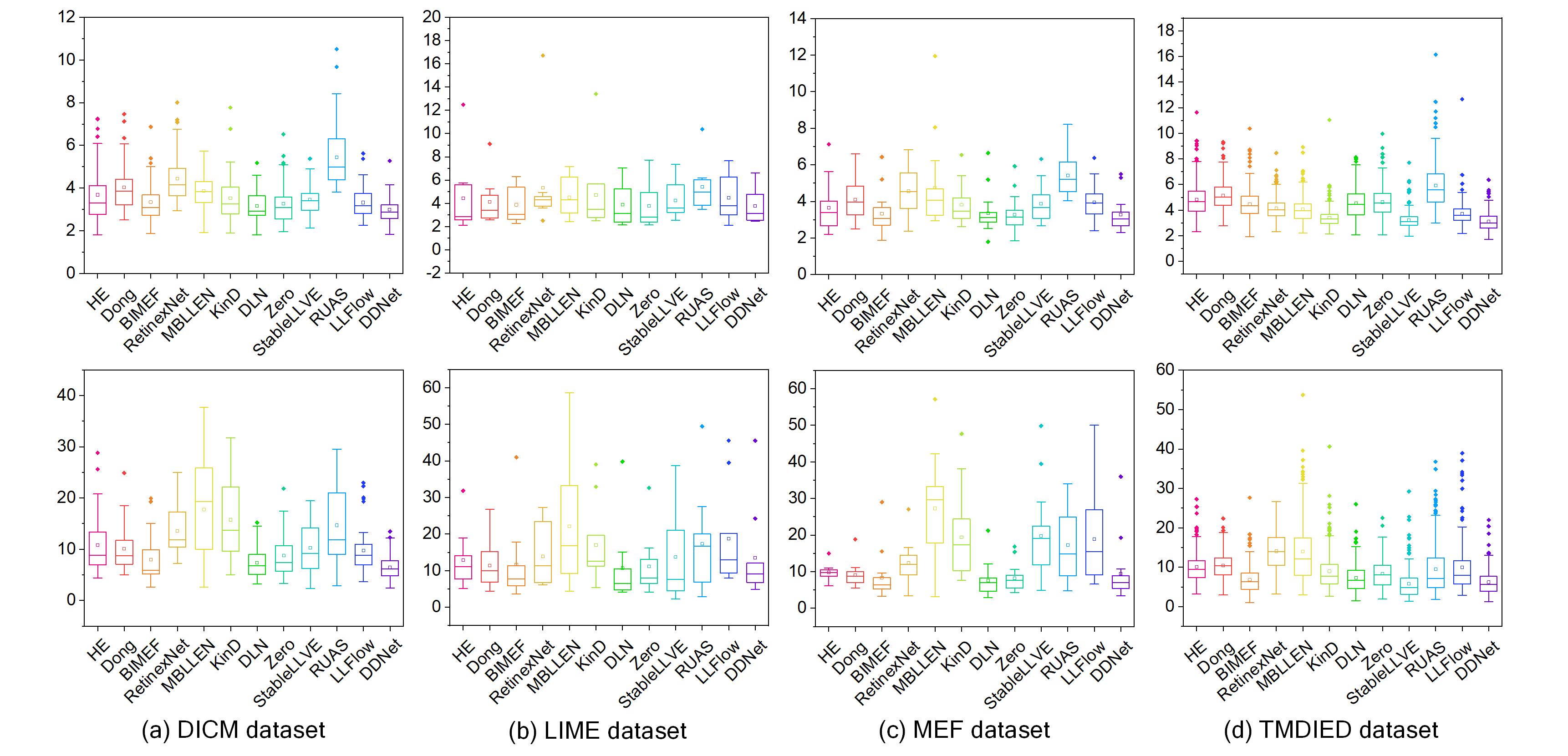}
    	\caption{The quantitative comparisons of enhancement methods on different datasets. From left to right: (a) DICM \cite{guo2020zero}, (b) LIME \cite{improving3}, (c) MEF \cite{lee2011power}, and (d) TMDIED datasets.  NIQE (top) and PIQE (bottom) are employed as the quantitative evaluation indices.}
    	\label{Figure_box}
    \end{figure*}
    \begin{figure*}[t]
    	\centering
    	\setlength{\abovecaptionskip}{0.1cm}
    	\includegraphics[width=1.0\linewidth]{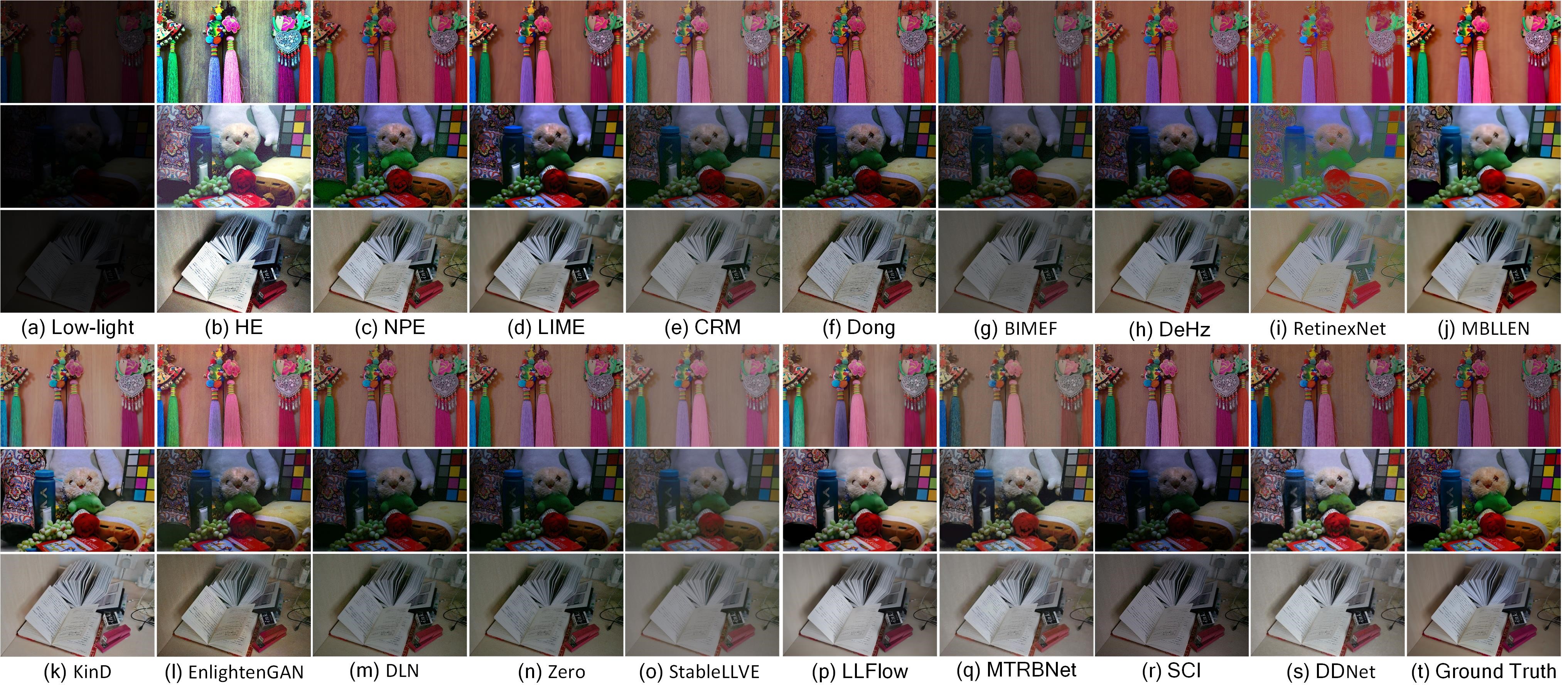}
    	\caption{The visual comparisons of different enhancement methods for three typical images from the LOL dataset \cite{wei2018deep}. From left to right: (a) Low-light images, restored images, generated by (b) HE \cite{HE}, (c) NPE \cite{improving1}, (d) LIME \cite{improving3}, (e) CRM \cite{ying2017new}, (f) Dong \cite{dong2010fast}, (g) BIMEF \cite{ying2017bio}, (h) DeHz \cite{jiang2013night}, (i) RetinexNet \cite{wei2018deep}, (j) MBLLEN \cite{lv2018mbllen}, (k) KinD \cite{zhang2019kindling}, (l) EnlightenGAN \cite{jiang2021enlightengan}, (m) DLN \cite{wang2020lightening}, (n) Zero \cite{guo2020zero}, (o) StableLLVE \cite{zhang2021learning}, (p) LLFlow \cite{wang2022low}, (q) MTRBNet \cite{lu2022mtrbnet}, (r) SCI \cite{ma2022toward}, (s) the proposed DDNet, and (t) Ground Truth, respectively.}
    	\label{Figure_LOL}
    \end{figure*}
    \setlength{\tabcolsep}{2.5pt}
    
\subsubsection{Evaluation Metrics}
    For low-light image enhancement, the evaluation metrics can be broadly classified into two groups: with or without the reference of ground truth. To conduct a more comprehensive analysis of the enhancement effectiveness, we first utilize the peak signal-to-noise ratio (PSNR) \cite{wang2009mean}, structural similarity (SSIM) \cite{wang2004image}, and learned perceptual image patch similarity (LPIPS) \cite{zhang2018unreasonable} as our reference-based evaluation metrics. Additionally, we have incorporated the natural image quality evaluator (NIQE) \cite{NIQE} and perceptual-based image quality evaluator (PIQE) \cite{venkatanath2015blind} as our no-reference metrics to quantitatively evaluate the performance of image enhancement across diverse low-light scenarios. It is noteworthy that larger values of PSNR and SSIM, as well as smaller values of NIQE, PIQE, and LPIPS, are indicative of better image quality.

\subsubsection{Running Platform}
    In the training period, the Adam optimizer is employed to suggest 100 epochs for training DDNet. The initial learning rate of the optimizer is 0.001, which is multiplied by 0.1 after every 20 epochs. Besides, the experimental network is trained and tested in a Python 3.7 environment using the PyTorch software package. The computational device is a PC with an AMD EPYC 7543 32-Core Processor CPU accelerated by an Nvidia A40 GPU, which has also been widely used in industrial-grade servers (e.g., Advantech SKY-6000 series and Thinkmate GPX servers). The proposed method could be thus easily extended to the higher-level visual task (e.g., vehicle detection and tracking) in ITS.

\subsection{Image Quality Assessment}
    To assess the quality of low-light image enhancement, we compare DDNet with several state-of-the-art methods, including HE \cite{HE}, NPE \cite{improving1}, LIME \cite{improving3}, JIEP \cite{cai2017joint}, CRM \cite{ying2017new}, Dong \cite{dong2010fast}, BIMEF \cite{ying2017bio}, DeHz \cite{jiang2013night}, RetinexNet \cite{wei2018deep}, MBLLEN \cite{lv2018mbllen}, KinD \cite{zhang2019kindling}, EnlightenGAN \cite{jiang2021enlightengan}, DLN \cite{wang2020lightening}, Zero \cite{guo2020zero}, StableLLVE \cite{zhang2021learning}, RUAS \cite{liu2021retinex}, LLFlow \cite{wang2022low}, MTRBNet \cite{lu2022mtrbnet}, and SCI \cite{ma2022toward}. It is noted that the parameters of each model are loaded from the corresponding official file of model weight.

\subsubsection{Quantitative Analysis}
     We first compute objective evaluation metrics (PSNR, SSIM, NIQE, PIQE, and LPIPS) for 15 LOL test images. As presented in Table \ref{Table02_LOL}, LIME outperforms the Retinex-based approach (i.e., NPE) overall, with credit to the noise reduction achieved by BM3D. Furthermore, CRM utilizes a camera response model, which is more effective in extracting information from low-light backgrounds. Zero yields unsatisfactory results in extremely low-light regions. Although DLN utilizes both local and global features of low-light images and exhibits better generalization capabilities, the enhancement effect still falls short.  Compared with the state-of-the-arts, our DDNet has an obvious advantage in the objective evaluation indicators with better stability, which is beneficial from the comprehensive guidance of both color and gradient domains.
    We also made an objective evaluation of images on other public datasets, including DICM \cite{guo2020zero}, LIME \cite{improving3}, MEF \cite{lee2011power}, and TMDIED, as illustrated in Tables \ref{table_niqe} and \ref{table_piqe}. Traditional methods are relatively uneven because they are challenging to deal with the nonuniform noise. The learning methods can receive satisfactory performance on both low-light enhancement and noise suppression, which thus performs better. In addition, due to the decomposition and reconstruction of double-domain features, DDNet can effectively recover the valuable information hidden in the dark with better robustness. Therefore, the enhanced image can better satisfy the complex transportation scenes and has the best quantitative evaluation metric. In Fig. \ref{Figure_box}, we present the quantitative evaluation results with the box plots. The first row is the NIQE evaluation results, and the second row is the PIQE evaluation results. The non-referenced metrics indicate that our method has better image quality compared with the state-of-the-arts.

\subsubsection{Visual Analysis}
    \begin{figure*}[t]
    	\centering
    	\setlength{\abovecaptionskip}{0.1cm}
    	\includegraphics[width=1\linewidth]{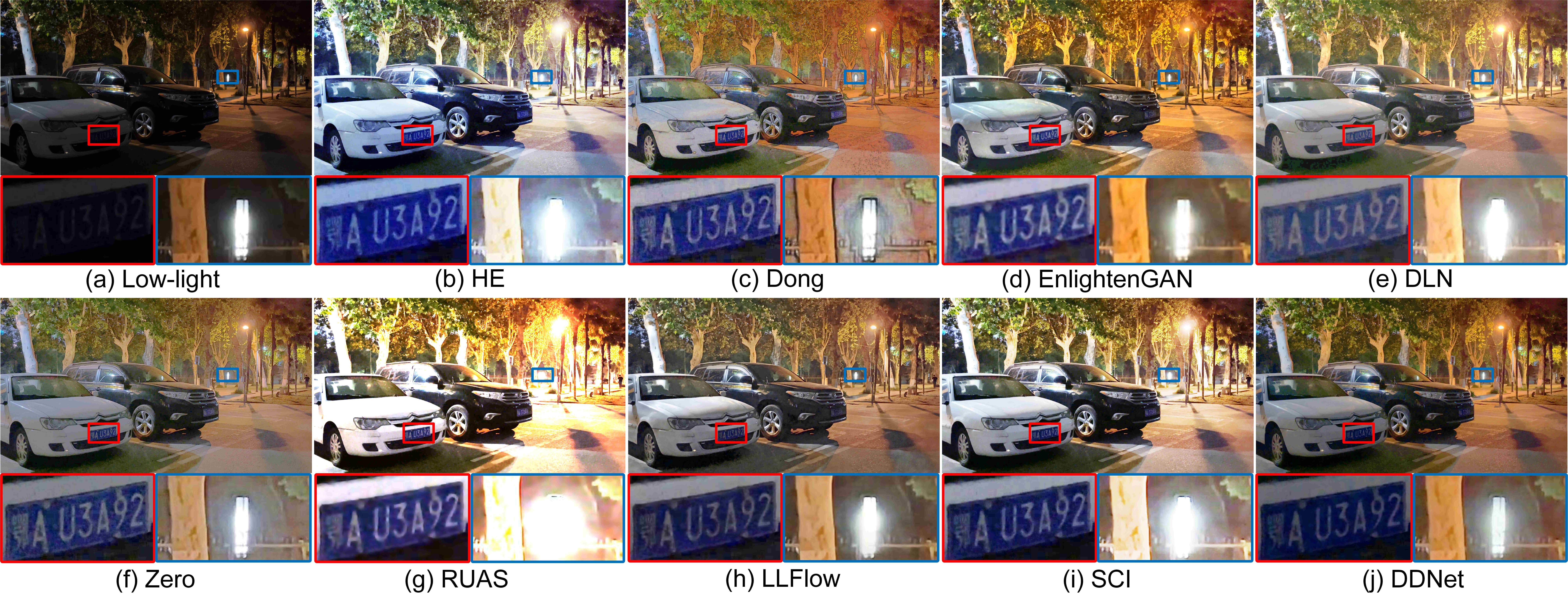}
    	\caption{The visual comparisons of different enhancement methods on the real-captured UHD low-light images in transportation surveillance. From left to right: (a) Low-light image, restored images generated by (b) HE \cite{HE}, (c) Dong \cite{dong2010fast}, (d) EnlightenGAN \cite{jiang2021enlightengan}, (e) DLN \cite{wang2020lightening}, (f) Zero \cite{guo2020zero}, (g) RUAS \cite{liu2021retinex}, (h) LLFlow \cite{wang2022low}, (i) SCI \cite{ma2022toward}, and (j) the proposed DDNet, respectively.}
    	\label{Figure_detail1}
    \end{figure*}
    \setlength{\tabcolsep}{5.0pt}
    \begin{table}[t]
    	\centering
    	\caption{{The ablation experiments on the SAM and SCM. The results are shown in PSNR, SSIM, and LPIPS on the 15 images from the LOL test dataset \cite{wei2018deep}. $\uparrow$ and $\downarrow$ represent that higher or lower values indicate better results, respectively.}}
    	\begin{tabular}{c|ccc}\hline
                Method     & PSNR $\uparrow$    & SSIM $\uparrow$   & LPIPS $\downarrow$ \\ \hline \hline
        w/o SAM, w/o SCM & 20.48$\pm$3.54 & 0.817$\pm$0.081 & 0.127$\pm$0.080       \\
        w SAM, w/o SCM   & 20.18$\pm$3.89 & 0.821$\pm$0.082 & 0.119$\pm$0.048       \\
        w/o SAM, w SCM   & 21.44$\pm$4.42 & 0.829$\pm$0.090 & 0.120$\pm$0.059       \\
        w SAM, w SCM     & 21.86$\pm$4.36 & 0.832$\pm$0.078 & 0.108$\pm$0.046      \\ \hline
        \end{tabular}\label{Table_Ablation}
    \end{table}
    \setlength{\tabcolsep}{5.0pt}
    \begin{table}[t]
		\centering
		\caption{{The ablation experiments on the GEM and CEM. The results are shown in PSNR, SSIM, and LPIPS on the 15 images from the LOL test dataset \cite{wei2018deep}. $\uparrow$ and $\downarrow$ represent that higher or lower values indicate better results, respectively.}}
        \begin{tabular}{c|ccc}
        \hline
                    Method     & PSNR $\uparrow$    & SSIM $\uparrow$   & LPIPS $\downarrow$ \\ \hline \hline
        w/o GEM, w/o CEM & 21.01$\pm$3.95 & 0.823$\pm$0.087 & 0.127$\pm$0.058          \\
        w GEM, w/o CEM   & 22.03$\pm$4.05 & 0.829$\pm$0.078 & 0.122$\pm$0.056          \\
        w/o GEM, w CEM   & 21.75$\pm$3.90 & 0.832$\pm$0.080 & 0.118$\pm$0.054          \\
        w GEM, w CEM     & 21.86$\pm$4.36 & 0.832$\pm$0.078 & 0.108$\pm$0.046          \\ \hline
        \end{tabular}
        \label{Table_Ablation2}
    \end{table}
    \begin{figure*}[t]
        \centering
        \setlength{\abovecaptionskip}{0.1cm}
        \includegraphics[width=1.0\linewidth]{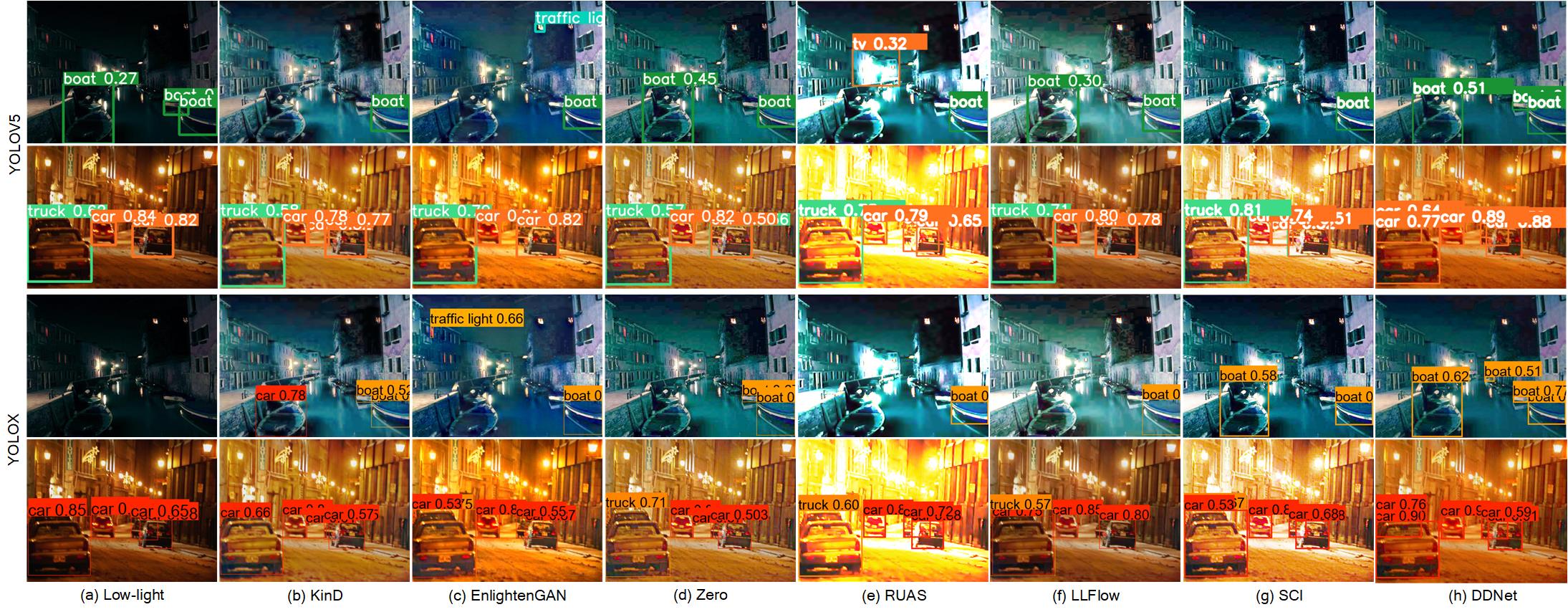}
        \caption{The qualitative results of object detection experiments on low-light transportation surveillance data, which select YOLOv5 and YOLOX \cite{ge2021yolox} as the basic detection methods. From left to right: (a) Low-light images, the enhanced images of (b) KinD \cite{zhang2019kindling}, (c) EnlightenGAN \cite{jiang2021enlightengan}, (d) Zero \cite{guo2020zero}, (e) RUAS \cite{liu2021retinex}, (f) LLFlow \cite{wang2022low}, (g) SCI \cite{ma2022toward}, and (h) the proposed DDNet, respectively. It can be seen that DDNet is more beneficial for detection accuracy improvement due to the enhancement of edge features on the gradient domain.}
        \label{detection}
    \end{figure*}
    \setlength{\tabcolsep}{4.0pt}
% Please add the following required packages to your document preamble:
% \usepackage{multirow}
    \begin{table}[t]
    \centering
        \caption{The ablation experiments on the weights of loss functions. The results are shown in PSNR, SSIM, and LPIPS on the 15 images from the LOL test dataset \cite{wei2018deep}. $\uparrow$ and $\downarrow$ represent that higher or lower values indicate better results, respectively. The best results are shown in \textbf{bold}}
        \label{tab:Ablation_loss}
    \begin{tabular}{cccc|cccl}
    \hline
    Class & $\omega_{1}$ & $\omega_{2}$ & $\omega_{3}$ & PSNR $\uparrow$ & SSIM $\uparrow$ & \multicolumn{2}{c}{LPIPS $\downarrow$} \\ \hline \hline
    \multirow{6}{*}{\begin{tabular}[c]{@{}c@{}}$\omega_{1} + \omega_{2}$ \\ / $\omega_{3}$\end{tabular}} & 0.05         & 0.05         & 0.90         & 21.09$\pm$4.37  & 0.826$\pm$0.078 & \multicolumn{2}{c}{0.113$\pm$0.047}    \\
    & 0.10         & 0.10         & 0.80         & 20.64$\pm$4.02  & 0.825$\pm$0.082 & \multicolumn{2}{c}{0.114$\pm$0.045}    \\
    & 0.15         & 0.15         & 0.70         & 21.22$\pm$3.57  & 0.836$\pm$0.073 & \multicolumn{2}{c}{0.109$\pm$0.048}    \\
    & 0.25         & 0.25         & 0.50         & 19.40$\pm$4.16  & 0.817$\pm$0.081 & \multicolumn{2}{c}{0.112$\pm$0.046}    \\
    & 0.30         & 0.30         & 0.40         & 20.95$\pm$4.15  & 0.825$\pm$0.077 & \multicolumn{2}{c}{0.112$\pm$0.050}    \\
    & \textbf{0.20}         & \textbf{0.20}         & \textbf{0.60}         &  \textbf{21.86$\pm$4.36}  & \textbf{0.832$\pm$0.078} & \multicolumn{2}{c}{\textbf{0.108$\pm$0.046}}    \\ \hline
    \multirow{4}{*}{$\omega_{1}$ / $\omega_{2}$}              & 0.05         & 0.35         & 0.60         & 21.45$\pm$4.63  & 0.823$\pm$0.086 & \multicolumn{2}{c}{0.118$\pm$0.053}    \\
    & 0.10         & 0.30         & 0.60         & 20.57$\pm$4.57  & 0.827$\pm$0.086 & \multicolumn{2}{c}{0.110$\pm$0.045}    \\
    & 0.30         & 0.10         & 0.60         & 20.54$\pm$4.67  & 0.826$\pm$0.088 & \multicolumn{2}{c}{0.113$\pm$0.045}    \\
    & \textbf{0.20}         & \textbf{0.20}         & \textbf{0.60}         & \textbf{21.86$\pm$4.36}  & \textbf{0.832$\pm$0.078} & \multicolumn{2}{c}{\textbf{0.108$\pm$0.046}}    \\ \hline 
    \end{tabular}
    \end{table}
    \setlength{\tabcolsep}{1.0pt}
    \begin{table}[t]
        \centering
        \caption{The comparison of running time (unit: second) between the DDNet and other low-light image enhancement methods.}
        \label{running_time}
        \begin{tabular}{l|ccccc}
        \hline
        \multirow{2}{*}{Method} & \multirow{2}{*}{Platform} & \multicolumn{4}{c}{Image size}             \\ \cline{3-6} 
                                &                           & 800×600 & 1080×720 & 2560×1440 & 3840×2160 \\ \hline \hline
        HE \cite{HE}                      & Matlab                    & 0.033   & 0.048    & 0.162     & 0.326     \\
        Dong \cite{dong2010fast}                    & Matlab                    & 0.161   & 0.256    & ---     & ---     \\ \hline
        EnlightenGAN \cite{jiang2021enlightengan}                      & Tensorflow                & 0.053   & 0.073    & 0.218     & 0.561     \\
        DLN \cite{wang2020lightening}                     & Pytorch                   & 0.007   & 0.011    & 0.018     & 0.028     \\
        Zero \cite{guo2020zero}                    & Pytorch                   & 0.001   & 0.001    & 0.001     & 0.001     \\
        RUAS \cite{liu2021retinex}                    & Pytorch                   & 0.006   & 0.010    & 0.023     & 0.030     \\
        LLFlow \cite{wang2022low}                  & Pytorch                   & 0.383   & 0.410    & ---     & ---     \\ 
         SCI \cite{ma2022toward}                  & Pytorch                   & 0.001   & 0.002    & 0.004     & 0.006     \\ \hline
        DDNet                   & Pytorch                   & 0.021   & 0.021    & 0.023     & 0.027     \\ \hline
        \end{tabular}
    \end{table}
    \begin{figure}[t]
    	\centering
    	\setlength{\abovecaptionskip}{0.1cm}
    	\includegraphics[width=1.0\linewidth]{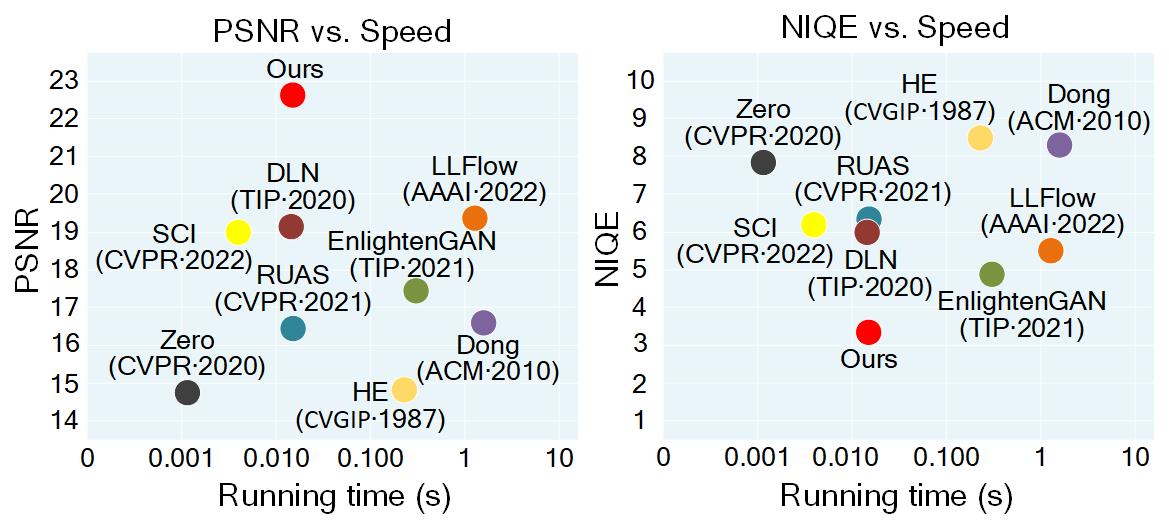}
    	\caption{The trade-off between the running time, NIQR, and PSNR on 4K images (3840 × 2160 pixels). The results show the superiority of our DDNet among the start-of-the-art methods.}
    	\label{Figure_running_time}
    \end{figure}

    To compare the visual performance of our DDNet with the state-of-the-arts, we first analyze the visual differences in the standard LOL test dataset. As shown in Fig. \ref{Figure_LOL}, HE has demonstrated significant improvements in the brightness and contrast of low-light images with rapid computational efficiency. However, it lacks the capability to suppress the noise and results in color distortion in local areas. NPE and BIMEF exhibit similar visual performance with poor contrast. Although LIME can eliminate noise in localized regions of the image, the BM3D algorithm struggles to distinguish between noise and texture information. CRM produces severely skewed color information in comparison to Retinex-based methods. RetinexNet demonstrates promising color extraction capabilities, but the edge feature is often severely compromised. MBLLEN and KinD can effectively remove unwanted noise information; however, the color naturalness is often unsatisfactory. EnlightenGAN, which employs a weakly-supervised architecture, can achieve low-light enhancement, but it is ineffective in extremely dark areas. Zero is lightweight and efficient, but the enhancement effect is often compromised for the sake of computational speed. DLN suffers from noise interference, which limits its effectiveness. While the StableLLVE recovers a significant amount of valuable information from dark regions, the resulting image is often overexposed, leading to a gray-and-white image with minimal contrast. SCI exhibits unsatisfactory performance when applied to extremely low-light images. By comparison, our proposed DDNet achieves a better balance between brightness enhancement and noise suppression in comparison to the current state-of-the-art methods. 
    %The color and edge features of the enhanced image are significantly restored, resulting in the superior image quality.
    %
    
    To verify the robustness of the proposed method on low-light transportation surveillance, we also collect UHD low-light images in transportation surveillance for testing\footnote{The real-captured UHD low-light images in transportation surveillance are available at: \url{https://github.com/QuJX/DDNet}.}. The comparison of global naturalness and local magnification of the enhanced image is shown in Fig. \ref{Figure_detail1}. HE, RUAS, and SCI suffer from overexposure, resulting in unnatural observation of luminous objects. Dong and Zero fail to satisfactorily recover the color features. EnlightenGAN and DLN are significantly interfered by the noise. LLFlow performs better, but the computational speed on 4K image can not meet the real-time video surveillance. In general, the double domain guided DDNet can achieve both satisfactory enhancement and computational efficiency.

\subsubsection{Running Time Comparisons}
    To prove the advantage of DDNet in terms of computational efficiency, we compare the performance on the running time with the objective indicators of the enhancement performance, as shown in Table. \ref{running_time} and Fig. \ref{Figure_running_time}. It is noted that the time over one second is shown in '---', which is not worth considering in UHD transportation surveillance due to the poor efficiency. With the outperforming enhancement performance, our method is able to enhance the 4K images over 35 FPS on the experimental platform, which is faster than most of the previous methods, meeting the requirements of UHD transportation surveillance. Although Zero \cite{guo2020zero} and SCI \cite{ma2022toward} are faster, their enhancement effect is much worse than ours. 

\subsection{Ablation Study}\label{Ablation Study}
    In this section, we attempt to verify the necessities of ScCAM and double-domain guidance. The 15 images from the LOL test dataset are utilized as the basic reference. According to the metrics provided in Table \ref{Table_Ablation}, the employment of the spatial attention module (SAM) and standard convolution module (SCM) significantly improves the enhancement performance. When both SAM and SCM are employed, PSNR, SSIM, and LPIPS performance are improved by $1.38$, $0.015$, and $0.019$, respectively. The experimental results about double-domain guidance are illustrated in Table \ref{Table_Ablation2}. The objective evaluation performance is the worst when the information of both color and gradient domains is not enhanced. The employment of coarse enhancement module (CEM) and LoG-based gradient enhancement module (GEM) significantly improves the enhancement performance. When both CEM and GEM are employed, PSNR, SSIM, and LPIPS performance are improved by $0.85$, $0.009$, and $0.019$, respectively.

    In addition, to verify the balance between the constraint on different domains, we conduct the ablation experiment on the design of loss function. Specifically, we set the weight of each loss differently in the training period. Table \ref{tab:Ablation_loss} presents the quantitative result. Firstly, we fix the weight ratio between $\omega_1$ and $\omega_2$ and adjust the ratio between them and $\omega_3$. We then fix $\omega_3$ as the obtained best result and adjust the ratio between $\omega_1$ and $\omega_2$. The ablation experiment indicates that current weights can supervise the network better with more satisfactory enhancement results.
    
    % %
    % \begin{figure*}[!ht]
    %     \centering
    %     \setlength{\abovecaptionskip}{0.1cm}
    %     \includegraphics[width=1.0\linewidth]{images/Figure_segmentation.jpg}
    %     \caption{\color{red}The results of segmentation experiments on synthetic low-light transportation-related visual data, which selects deeplabv3+ \cite{chen2018encoder} as the basic segmentation method. The first and third rows are visible light images, and the second and fourth rows are the results of scene segmentation. From left-top to right-bottom: (a) Low-light image, the enhanced images of (b) RetinexNet \cite{wei2018deep}, (c) KinD \cite{zhang2019kindling}, (d) EnlightenGAN \cite{jiang2021enlightengan}, (e) Zero \cite{guo2020zero}, (f) RUAS \cite{liu2021retinex}, (g) LLFlow \cite{wang2022low}, (h) SCI \cite{ma2022toward}, (i) the proposed DDNet, and (j) Ground Truth, respectively. Due to the enhancement on both color and gradient domains, the edge features are recovered better in DDNet, which enables the segmentation network to classify the farther small vehicles with better accuracy.}
    %     \label{segmentation}
    % \end{figure*}
    %

    %
    \begin{figure*}[!ht]
        \centering
        \setlength{\abovecaptionskip}{0.1cm}
        \includegraphics[width=1.0\linewidth]{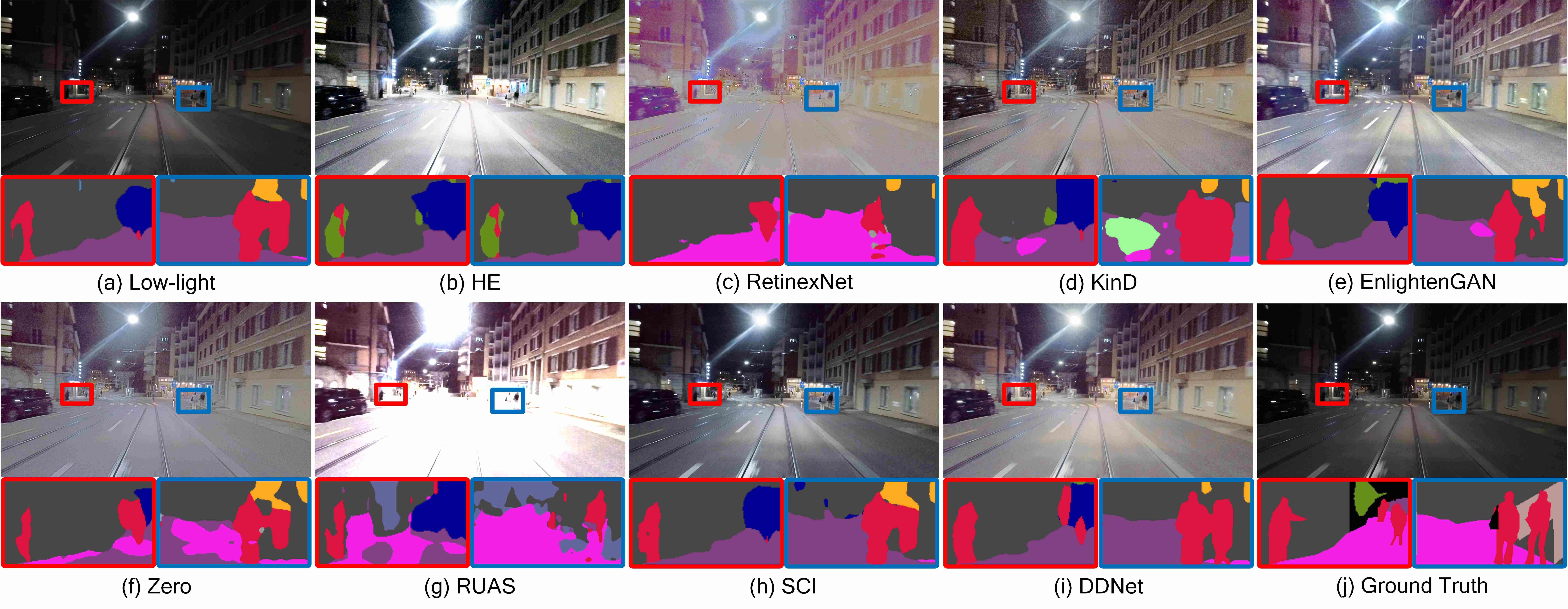}
        \caption{The detailed results of segmentation experiments on the ACDC dataset \cite{sakaridis2021acdc}, which selects DAFormer \cite{hoyer2022daformer} as the basic segmentation method. The first and third rows are raw images, and the second and fourth rows are the visualized results of scene segmentation. From left-top to right-bottom: (a) Low-light image, and the segmentation results on the enhanced images of (b) HE \cite{HE}, (c) RetinexNet \cite{wei2018deep}, (d) KinD \cite{zhang2019kindling}, (e) EnlightenGAN \cite{jiang2021enlightengan}, (f) Zero \cite{guo2020zero}, (g) RUAS \cite{liu2021retinex}, (h) SCI \cite{ma2022toward}, (i) the proposed DDNet, and (j) Ground Truth, respectively. It is noted that the employed DAFormer is pre-trained on cityscapes dataset. Compared with other methods, our DDNet enables the model pre-trained on normal-light images performing better under low-light conditions.}
        \label{segmentation_ACDC}
    \end{figure*}

\subsection{Improvement of Object Detection in ITS}
    In order to further demonstrate the practical benefits of our proposed DDNet in the domain of transportation surveillance, we have employed the YOLOv5 and YOLOX \cite{ge2021yolox} to detect objects under low-light conditions, and compare the detection results with or without the application of image enhancement methods. To conduct our analysis, we have selected experimental images from the COCO \cite{lin2014microsoft} and ExDARK \cite{loh2019getting} datasets. Specifically, we initially selected 1500 transportation-related images from the COCO dataset for the training of our detection networks. Subsequently, we performed evaluation tests on the ExDark dataset. As depicted in Fig. \ref{detection}, the detection networks exhibit poor performance in dark transportation scenes, often failing to achieve accurate object detection owing to the low contrast and vague edge features. However, following the application of enhancement methods, the detection accuracy is significantly increased. Furthermore, in comparison to state-of-the-art methods, the images enhanced by DDNet demonstrate superior performance, primarily due to the comprehensive recovery of both color and gradient features. These findings provide the evidence that DDNet holds practical benefits for low-light transportation surveillance tasks, and is beneficial for higher-level visual tasks in ITS when operating under low-light environments.

\subsection{Improvement of Scene Segmentation in ITS}
    The scene segmentation is also a typical higher-lever visual task in transportation surveillance. To demonstrate the practical improvement of our method for scene segmentation, we conducted the comparison experiment on ACDC \cite{sakaridis2021acdc}, a real-captured transportation-related dataset under adverse visual conditions, including low-light, hazy, rainy, etc. We employed the DAFormer \cite{hoyer2022daformer} with the model weight pre-trained on cityscapes dataset, which mainly consists of normal-light images. Fig. \ref{segmentation_ACDC} presents the visual results. As can be observed, in low-light environments, the edge features of objects appear vague, and the color brightness is low, making it challenging for segmentation methods to accurately classify the pixels. Additionally, accurately classifying small objects, such as distant pedestrians, is difficult owing to the low contrast. Following the application of low-light image enhancement method, the visibility of low-light scenes is significantly improved. However, most state-of-the-art methods tend to suffer from noise interference and color distortion, leading to erroneous segmentation. Furthermore, it is still challenging to accurately segment small objects due to the vague edge features. In particular, our DDNet effectively recovers the low-light image with better color naturalness and clear edge features, resulting in more accurate classification of challenging pixels in the enhanced images. Overall, our method enables models pre-trained on normal-light images to perform better in low-light conditions.
    %{\color{red} Specifically, for the synthesized low-light images, we select Deeplabv3+ \cite{chen2018encoder} as the basic segmentation method.} The clear experimental image is selected from the cityscapes dataset \cite{cordts2016cityscapes}, and the low-light images are generated via Eq. (\ref{synthetic}). The pre-trained model of the experimental deeplabv3+ is downloaded from the official website. As shown in Fig. \ref{segmentation}, in the low-light environment, the edge features of the objects are vague, and the color brightness is low, which makes the segmentation method difficult to classify the pixels accurately. The small objects, like distant cars, are challenging to segment due to the low contrast. After the enhancement, the visibility under low-light conditions is greatly improved. However, most state-of-the-art methods suffer from noise interference and color distortion, which makes the method operate mistaken segmentation. Besides, small objects still can not be accurately segmented for vague edge features. After the enhancement of DDNet, the low-light image is finely recovered with better color naturalness and clear edge features. The challenging pixels are thus classified more accurately in the enhanced images. 
    
\section{Conclusion and Future Perspectives}\label{Conclusions and Discussion}
    This paper proposes a double domain guided real-time low-light image enhancement network (DDNet) for UHD transportation surveillance. Specifically, we suggest the encoder-decoder structure as the main architecture of the learning network, and the original task is divided into two subtasks (i.e., coarse enhancement and Laplacian of Gaussian (LoG)-based gradient enhancement). The coarse enhancement module (CEM) and LoG-based gradient enhancement module (GEM) are proposed and embedded in the encoder-decoder structure, which assist the network to efficiently enhance the color and gradient features under the constraint of the proposed joint loss function. Through the decomposition and reconstruction of both color and gradient features, our DDNet can perceive the detailed information concealed by the dark background with greater precision. Image quality and running time experiments on standard datasets and UHD low-light images in transportation surveillance demonstrate that our DDNet satisfies the requirement of real-time transportation surveillance. Besides, compared with the state-of-the-arts, the object detection and segmentation experiments prove that our method contributes more to higher-level image analysis tasks under low-light environments in ITS. It is mainly  beneficial from the guidance of both color and gradient domains. 

    In conclusion, our work presents a real-time low-light image enhancement method for UHD transportation surveillance in ITS. Although our method obtains promising results in this study, it still faces several challenges, e.g., inadequate real-captured dataset and relative large model size. The further improvement of our method includes follows.
      \begin{itemize}
          \item To overcome the inadequate real-captured dataset, the semi-supervised architecture and generative adversarial networks (GAN) will be considered to reduce the dependence of our DDNet on paired datasets.
          \item Currently, although the proposed method achieves the real-time processing for transportation surveillance, the model size is not lightweight enough. In the future, we will consider to employ the pruning technology \cite{liu2018rethinking} to build more lightweight models. 
          \item To overcome the blurred appearance features of the fast-moving objects in real-time transportation surveillance (e.g., the vehicles on the expressways), we will consider to utilize the multi-task learning to achieve image deblurring and enhancement simultaneously.
      \end{itemize}

\bibliography{Reference}    

% Generated by IEEEtran.bst, version: 1.14 (2015/08/26)
\begin{thebibliography}{10}
\providecommand{\url}[1]{#1}
\csname url@samestyle\endcsname
\providecommand{\newblock}{\relax}
\providecommand{\bibinfo}[2]{#2}
\providecommand{\BIBentrySTDinterwordspacing}{\spaceskip=0pt\relax}
\providecommand{\BIBentryALTinterwordstretchfactor}{4}
\providecommand{\BIBentryALTinterwordspacing}{\spaceskip=\fontdimen2\font plus
\BIBentryALTinterwordstretchfactor\fontdimen3\font minus \fontdimen4\font\relax}
\providecommand{\BIBforeignlanguage}[2]{{%
\expandafter\ifx\csname l@#1\endcsname\relax
\typeout{** WARNING: IEEEtran.bst: No hyphenation pattern has been}%
\typeout{** loaded for the language `#1'. Using the pattern for}%
\typeout{** the default language instead.}%
\else
\language=\csname l@#1\endcsname
\fi
#2}}
\providecommand{\BIBdecl}{\relax}
\BIBdecl

\bibitem{jiang2022unsupervised}
Q.~Jiang, Y.~Mao, R.~Cong, W.~Ren, C.~Huang, and F.~Shao, ``Unsupervised decomposition and correction network for low-light image enhancement,'' \emph{IEEE Trans. Intell. Transp. Syst.}, vol.~23, no.~10, pp. 19\,440--19\,455, Apr. 2022.

\bibitem{liang2022edge}
S.~Liang, H.~Wu, L.~Zhen, Q.~Hua, S.~Garg, G.~Kaddoum, M.~M. Hassan, and K.~Yu, ``Edge yolo: Real-time intelligent object detection system based on edge-cloud cooperation in autonomous vehicles,'' \emph{IEEE Trans. Intell. Transp. Syst.}, vol.~23, no.~12, pp. 25\,345--25\,360, Mar. 2022.

\bibitem{ge2021yolox}
Z.~Ge, S.~Liu, F.~Wang, Z.~Li, and J.~Sun, ``Yolox: Exceeding yolo series in 2021,'' \emph{arXiv preprint arXiv:2107.08430}, Aug. 2021.

\bibitem{qin2022id}
L.~Qin, Y.~Shi, Y.~He, J.~Zhang, X.~Zhang, Y.~Li, T.~Deng, and H.~Yan, ``Id-yolo: Real-time salient object detection based on the driver's fixation region,'' \emph{IEEE Trans. Intell. Transp. Syst.}, vol.~23, no.~9, pp. 15\,898--15\,908, Sep. 2022.

\bibitem{gao2021mscfnet}
G.~Gao, G.~Xu, Y.~Yu, J.~Xie, J.~Yang, and D.~Yue, ``Mscfnet: a lightweight network with multi-scale context fusion for real-time semantic segmentation,'' \emph{IEEE Trans. Intell. Transp. Syst.}, vol.~23, no.~12, pp. 25\,489--25\,499, Aug. 2021.

\bibitem{zhang2022trans4trans}
J.~Zhang, K.~Yang, A.~Constantinescu, K.~Peng, K.~M{\"u}ller, and R.~Stiefelhagen, ``Trans4trans: Efficient transformer for transparent object and semantic scene segmentation in real-world navigation assistance,'' \emph{IEEE Trans. Intell. Transp. Syst.}, vol.~23, no.~10, pp. 19\,173--19\,186, Oct. 2022.

\bibitem{weng2022deep}
X.~Weng, Y.~Yan, G.~Dong, C.~Shu, B.~Wang, H.~Wang, and J.~Zhang, ``Deep multi-branch aggregation network for real-time semantic segmentation in street scenes,'' \emph{IEEE Trans. Intell. Transp. Syst.}, vol.~23, no.~10, pp. 17\,224--17\,240, Oct. 2022.

\bibitem{krishnan2009dark}
D.~Krishnan and R.~Fergus, ``Dark flash photography,'' \emph{ACM Trans. Graph.}, vol.~28, no.~3, p.~96, 2009.

\bibitem{shen2022parkpredict+}
X.~Shen, M.~Lacayo, N.~Guggilla, and F.~Borrelli, ``Parkpredict+: Multimodal intent and motion prediction for vehicles in parking lots with cnn and transformer,'' in \emph{Proc. IEEE ITSC}, 2022, pp. 3999--4004.

\bibitem{kujawski2019concept}
A.~Kujawski, J.~Lemke, and T.~Dudek, ``Concept of using unmanned aerial vehicle (uav) in the analysis of traffic parameters on oder waterway,'' \emph{Transp. Res. Rec.}, vol.~39, pp. 231--241, 2019.

\bibitem{wang2022improved}
K.~Wang, T.~Yang, X.~Zhang, and M.~Tang, ``An improved ultra-high definition panoramic video mosaic method for airport scene,'' in \emph{Proc. IEEE ICCC}, 2022, pp. 2076--2080.

\bibitem{lin2022uhd}
Q.~Lin, Z.~Zheng, and X.~Jia, ``Uhd low-light image enhancement via interpretable bilateral learning,'' \emph{Inf. Sci.}, vol. 608, pp. 1401--1415, Aug. 2022.

\bibitem{zhang2021night}
X.~Zhang, B.~Story, and D.~Rajan, ``Night time vehicle detection and tracking by fusing vehicle parts from multiple cameras,'' \emph{IEEE Trans. Intell. Transp. Syst.}, vol.~23, no.~7, pp. 8136--8156, May 2021.

\bibitem{yang2020part}
P.~Yang, G.~Zhang, L.~Wang, L.~Xu, Q.~Deng, and M.-H. Yang, ``A part-aware multi-scale fully convolutional network for pedestrian detection,'' \emph{IEEE Trans. Intell. Transp. Syst.}, vol.~22, no.~2, pp. 1125--1137, Jan. 2020.

\bibitem{li2020highly}
Q.~Li, S.~Garg, J.~Nie, X.~Li, R.~W. Liu, Z.~Cao, and M.~S. Hossain, ``A highly efficient vehicle taillight detection approach based on deep learning,'' \emph{IEEE Trans. Intell. Transp. Syst.}, vol.~22, no.~7, pp. 4716--4726, Oct. 2020.

\bibitem{HE}
S.~M. Pizer, E.~P. Amburn, J.~D. Austin, R.~Cromartie, A.~Geselowitz, T.~Greer, B.~ter Haar~Romeny, J.~B. Zimmerman, and K.~Zuiderveld, ``Adaptive histogram equalization and its variations,'' \emph{Comput. Vis., Graph., Image Process.}, vol.~39, no.~3, pp. 355--368, Oct. 1987.

\bibitem{dhara2021exposedness}
S.~K. Dhara and D.~Sen, ``Exposedness-based noise-suppressing low-light image enhancement,'' \emph{IEEE Trans. Circuits Syst. Video Technol.}, vol.~32, no.~6, pp. 3438--3451, Sep. 2021.

\bibitem{wang2022multiple}
Z.~Wang, Z.~Li, J.~Leng, M.~Li, and L.~Bai, ``Multiple pedestrian tracking with graph attention map on urban road scene,'' \emph{IEEE Trans. Intell. Transp. Syst.}, vol.~24, pp. 8567--8579, Aug. 2022.

\bibitem{yang2021ndnet}
Z.~Yang, H.~Yu, Q.~Fu, W.~Sun, W.~Jia, M.~Sun, and Z.-H. Mao, ``Ndnet: Narrow while deep network for real-time semantic segmentation,'' \emph{IEEE Trans. Intell. Transp. Syst.}, vol.~22, no.~9, pp. 5508--5519, Apr. 2021.

\bibitem{xiong2020vehicle}
Z.~Xiong, M.~Li, Y.~Ma, and X.~Wu, ``Vehicle re-identification with image processing and car-following model using multiple surveillance cameras from urban arterials,'' \emph{IEEE Trans. Intell. Transp. Syst.}, vol.~22, no.~12, pp. 7619--7630, Jul. 2020.

\bibitem{kumar2020efficient}
R.~Kumar, R.~Balasubramanian, and B.~K. Kaushik, ``Efficient method and architecture for real-time video defogging,'' \emph{IEEE Trans. Intell. Transp. Syst.}, vol.~22, no.~10, pp. 6536--6546, Oct. 2020.

\bibitem{guo2020zero}
C.~Guo, C.~Li, J.~Guo, C.~C. Loy, J.~Hou, S.~Kwong, and R.~Cong, ``Zero-reference deep curve estimation for low-light image enhancement,'' in \emph{Proc. IEEE CVPR}, 2020, pp. 1780--1789.

\bibitem{ma2022toward}
L.~Ma, T.~Ma, R.~Liu, X.~Fan, and Z.~Luo, ``Toward fast, flexible, and robust low-light image enhancement,'' in \emph{Proc. IEEE CVPR}, 2022, pp. 5637--5646.

\bibitem{li2023embedding}
C.~Li, C.-L. Guo, M.~Zhou, Z.~Liang, S.~Zhou, R.~Feng, and C.~C. Loy, ``Embedding fourier for ultra-high-definition low-light image enhancement,'' \emph{arXiv preprint arXiv:2302.11831}, Feb. 2023.

\bibitem{retinex}
E.~H. Land, ``The retinex theory of color vision,'' \emph{Sci. Am.}, vol. 237, no.~6, pp. 108--129, Dec. 1977.

\bibitem{improving1}
S.~Wang, J.~Zheng, H.-M. Hu, and B.~Li, ``Naturalness preserved enhancement algorithm for non-uniform illumination images,'' \emph{IEEE Trans. Image Process.}, vol.~22, no.~9, pp. 3538--3548, May 2013.

\bibitem{improving3}
X.~Guo, Y.~Li, and H.~Ling, ``Lime: Low-light image enhancement via illumination map estimation,'' \emph{IEEE Trans. Image Process.}, vol.~26, no.~2, pp. 982--993, Dec. 2016.

\bibitem{cai2017joint}
B.~Cai, X.~Xu, K.~Guo, K.~Jia, B.~Hu, and D.~Tao, ``A joint intrinsic-extrinsic prior model for retinex,'' in \emph{Proc. IEEE ICCV}, 2017, pp. 4000--4009.

\bibitem{ying2017new}
Z.~Ying, G.~Li, Y.~Ren, R.~Wang, and W.~Wang, ``A new low-light image enhancement algorithm using camera response model,'' in \emph{Proc. IEEE ICCV}, 2017, pp. 3015--3022.

\bibitem{ying2017bio}
Z.~Ying, G.~Li, and W.~Gao, ``A bio-inspired multi-exposure fusion framework for low-light image enhancement,'' \emph{arXiv preprint arXiv:1711.00591}, Nov. 2017.

\bibitem{dong2010fast}
X.~Dong, Y.~Pang, and J.~Wen, ``Fast efficient algorithm for enhancement of low lighting video,'' in \emph{ACM SIGGRAPH 2010 Posters}, Jul. 2010, pp. 1--1.

\bibitem{jiang2013night}
X.~Jiang, H.~Yao, S.~Zhang, X.~Lu, and W.~Zeng, ``Night video enhancement using improved dark channel prior,'' in \emph{Proc. IEEE ICIP}, 2013, pp. 553--557.

\bibitem{9743313}
L.~Ma, R.~Liu, Y.~Wang, X.~Fan, and Z.~Luo, ``Low-light image enhancement via self-reinforced retinex projection model,'' \emph{IEEE Trans. Multimedia}, vol.~25, pp. 3573--3586, Mar. 2023.

\bibitem{lecun2015deep}
Y.~LeCun, Y.~Bengio, and G.~Hinton, ``Deep learning,'' \emph{NATURE}, vol. 521, no. 7553, pp. 436--444, May 2015.

\bibitem{zhang2019kindling}
Y.~Zhang, J.~Zhang, and X.~Guo, ``Kindling the darkness: A practical low-light image enhancer,'' in \emph{Proc. ACM MM}, 2019, pp. 1632--1640.

\bibitem{wei2018deep}
C.~Wei, W.~Wang, W.~Yang, and J.~Liu, ``Deep retinex decomposition for low-light enhancement,'' \emph{arXiv preprint arXiv:1808.04560}, Aug. 2018.

\bibitem{liu2021retinex}
R.~Liu, L.~Ma, J.~Zhang, X.~Fan, and Z.~Luo, ``Retinex-inspired unrolling with cooperative prior architecture search for low-light image enhancement,'' in \emph{Proc. IEEE CVPR}, 2021, pp. 10\,561--10\,570.

\bibitem{wu2022uretinex}
W.~Wu, J.~Weng, P.~Zhang, X.~Wang, W.~Yang, and J.~Jiang, ``Uretinex-net: Retinex-based deep unfolding network for low-light image enhancement,'' in \emph{Proc. IEEE CVPR}, 2022, pp. 5901--5910.

\bibitem{9056796}
X.~Ren, W.~Yang, W.-H. Cheng, and J.~Liu, ``Lr3m: Robust low-light enhancement via low-rank regularized retinex model,'' \emph{IEEE Trans. Image Process.}, vol.~29, pp. 5862--5876, Apr. 2020.

\bibitem{lv2018mbllen}
F.~Lv, F.~Lu, J.~Wu, and C.~Lim, ``Mbllen: Low-light image/video enhancement using cnns.'' in \emph{Proc. BMVC}, 2018, p.~4.

\bibitem{wang2020lightening}
L.-W. Wang, Z.-S. Liu, W.-C. Siu, and D.~P. Lun, ``Lightening network for low-light image enhancement,'' \emph{IEEE Trans. Image Process.}, vol.~29, pp. 7984--7996, Jul. 2020.

\bibitem{zhang2021learning}
F.~Zhang, Y.~Li, S.~You, and Y.~Fu, ``Learning temporal consistency for low light video enhancement from single images,'' in \emph{Proc. IEEE CVPR}, 2021, pp. 4967--4976.

\bibitem{wang2022low}
Y.~Wang, R.~Wan, W.~Yang, H.~Li, L.-P. Chau, and A.~Kot, ``Low-light image enhancement with normalizing flow,'' in \emph{Proc. AAAI}, 2022, pp. 2604--2612.

\bibitem{lu2022mtrbnet}
Y.~Lu, Y.~Guo, R.~W. Liu, and W.~Ren, ``Mtrbnet: Multi-branch topology residual block-based network for low-light enhancement,'' \emph{IEEE Signal Process. Lett.}, vol.~29, pp. 1127--1131, Mar. 2022.

\bibitem{guo2023low}
X.~Guo and Q.~Hu, ``Low-light image enhancement via breaking down the darkness,'' \emph{Int. J. Comput. Vis.}, vol. 131, no.~1, pp. 48--66, Oct. 2023.

\bibitem{jiang2021enlightengan}
Y.~Jiang, X.~Gong, D.~Liu, Y.~Cheng, C.~Fang, X.~Shen, J.~Yang, P.~Zhou, and Z.~Wang, ``Enlightengan: Deep light enhancement without paired supervision,'' \emph{IEEE Trans. Image Process.}, vol.~30, pp. 2340--2349, Jan. 2021.

\bibitem{yang2021band}
W.~Yang, S.~Wang, Y.~Fang, Y.~Wang, and J.~Liu, ``Band representation-based semi-supervised low-light image enhancement: Bridging the gap between signal fidelity and perceptual quality,'' \emph{IEEE Trans. Image Process.}, vol.~30, pp. 3461--3473, Mar. 2021.

\bibitem{wang2023ultra}
T.~Wang, K.~Zhang, T.~Shen, W.~Luo, B.~Stenger, and T.~Lu, ``Ultra-high-definition low-light image enhancement: a benchmark and transformer-based method,'' in \emph{Proc. AAAI}, vol.~37, no.~3, 2023, pp. 2654--2662.

\bibitem{lu2022low}
Y.~Lu, Y.~Gao, Y.~Guo, W.~Xu, and X.~Hu, ``Low-light image enhancement via gradient prior-aided network,'' \emph{IEEE Access}, vol.~10, pp. 92\,583--92\,596, Aug. 2022.

\bibitem{zheng2021image}
S.~Zheng and S.~Dai, ``Image enhancement for railway inspections based on cyclegan under the retinex theory,'' in \emph{Proc. IEEE ITSC}, 2021, pp. 2330--2335.

\bibitem{liu2022attention}
R.~W. Liu, N.~Liu, Y.~Huang, and Y.~Guo, ``Attention-guided lightweight generative adversarial network for low-light image enhancement in maritime video surveillance,'' \emph{J. Navig.}, vol.~75, no.~5, p. 1100–1117, Aug. 2022.

\bibitem{guo2022lightweight}
Y.~Guo, Y.~Lu, and R.~W. Liu, ``Lightweight deep network-enabled real-time low-visibility enhancement for promoting vessel detection in maritime video surveillance,'' \emph{J. Navig.}, vol.~75, no.~1, pp. 230--250, Oct. 2022.

\bibitem{liu2021enhanced}
R.~W. Liu, W.~Yuan, X.~Chen, and Y.~Lu, ``An enhanced cnn-enabled learning method for promoting ship detection in maritime surveillance system,'' \emph{Ocean Eng.}, vol. 235, p. 109435, Sep. 2021.

\bibitem{wu2022edge}
Y.~Wu, H.~Guo, C.~Chakraborty, M.~Khosravi, S.~Berretti, and S.~Wan, ``Edge computing driven low-light image dynamic enhancement for object detection,'' \emph{IEEE Trans. Network Sci. Eng.}, to be published, doi:{\color{blue}\href{http://dx.doi.org/10.1109/TNSE.2022.3151502}{10.1109/TNSE.2022.3151502}}.

\bibitem{chen2021high}
X.~Chen, Z.~Li, Y.~Yang, L.~Qi, and R.~Ke, ``High-resolution vehicle trajectory extraction and denoising from aerial videos,'' \emph{IEEE Trans. Intell. Transp. Syst.}, vol.~22, no.~5, pp. 3190--3202, May 2021.

\bibitem{han2021using}
H.-Y. Han, Y.-C. Chen, P.-Y. Hsiao, and L.-C. Fu, ``Using channel-wise attention for deep cnn based real-time semantic segmentation with class-aware edge information,'' \emph{IEEE Trans. Intell. Transp. Syst.}, vol.~22, no.~2, pp. 1041--1051, Feb. 2021.

\bibitem{wang2007laplacian}
X.~Wang, ``Laplacian operator-based edge detectors,'' \emph{IEEE Trans. Pattern Anal. Mach. Intell.}, vol.~29, no.~5, pp. 886--890, Mar. 2007.

\bibitem{zou2022self}
W.~Zou, T.~Ye, W.~Zheng, Y.~Zhang, L.~Chen, and Y.~Wu, ``Self-calibrated efficient transformer for lightweight super-resolution,'' in \emph{Proc. IEEE CVPR}, 2022, pp. 930--939.

\bibitem{wang2004image}
Z.~Wang, A.~C. Bovik, H.~R. Sheikh, and E.~P. Simoncelli, ``Image quality assessment: from error visibility to structural similarity,'' \emph{IEEE Trans. Image Process.}, vol.~13, no.~4, pp. 600--612, Apr. 2004.

\bibitem{everingham2010pascal}
M.~Everingham, L.~Van~Gool, C.~K. Williams, J.~Winn, and A.~Zisserman, ``The pascal visual object classes (voc) challenge,'' \emph{Int. J. Comput. Vis.}, vol.~88, no.~2, pp. 303--338, Sep. 2009.

\bibitem{lin2014microsoft}
T.-Y. Lin, M.~Maire, S.~Belongie, J.~Hays, P.~Perona, D.~Ramanan, P.~Doll{\'a}r, and C.~L. Zitnick, ``Microsoft coco: Common objects in context,'' in \emph{Proc. ECCV}, 2014, pp. 740--755.

\bibitem{wen2020ua}
L.~Wen, D.~Du, Z.~Cai, Z.~Lei, M.-C. Chang, H.~Qi, J.~Lim, M.-H. Yang, and S.~Lyu, ``Ua-detrac: A new benchmark and protocol for multi-object detection and tracking,'' \emph{Comput. Vision Image Understanding}, vol. 193, p. 102907, Apr. 2020.

\bibitem{lee2011power}
C.~Lee, C.~Lee, Y.-Y. Lee, and C.-S. Kim, ``Power-constrained contrast enhancement for emissive displays based on histogram equalization,'' \emph{IEEE Trans. Image Process.}, vol.~21, no.~1, pp. 80--93, Jun. 2011.

\bibitem{wang2009mean}
Z.~Wang and A.~C. Bovik, ``Mean squared error: Love it or leave it? a new look at signal fidelity measures,'' \emph{IEEE Signal Process Mag.}, vol.~26, no.~1, pp. 98--117, Jan. 2009.

\bibitem{zhang2018unreasonable}
R.~Zhang, P.~Isola, A.~A. Efros, E.~Shechtman, and O.~Wang, ``The unreasonable effectiveness of deep features as a perceptual metric,'' in \emph{Proc. IEEE CVPR}, 2018, pp. 586--595.

\bibitem{NIQE}
H.~Yeganeh and Z.~Wang, ``Objective quality assessment of tone-mapped images,'' \emph{IEEE Trans. Image Process.}, vol.~22, no.~2, pp. 657--667, Oct. 2012.

\bibitem{venkatanath2015blind}
N.~Venkatanath, D.~Praneeth, M.~C. Bh, S.~S. Channappayya, and S.~S. Medasani, ``Blind image quality evaluation using perception based features,'' in \emph{Proc. IEEE NCC}, 2015, pp. 1--6.

\bibitem{sakaridis2021acdc}
C.~Sakaridis, D.~Dai, and L.~Van~Gool, ``Acdc: The adverse conditions dataset with correspondences for semantic driving scene understanding,'' in \emph{Proc. IEEE ICCV}, 2021, pp. 10\,765--10\,775.

\bibitem{hoyer2022daformer}
L.~Hoyer, D.~Dai, and L.~Van~Gool, ``Daformer: Improving network architectures and training strategies for domain-adaptive semantic segmentation,'' in \emph{Proc. IEEE CVPR}, 2022, pp. 9924--9935.

\bibitem{loh2019getting}
Y.~P. Loh and C.~S. Chan, ``Getting to know low-light images with the exclusively dark dataset,'' \emph{Comput. Vision Image Understanding}, vol. 178, pp. 30--42, Jan. 2019.

\bibitem{liu2018rethinking}
Z.~Liu, M.~Sun, T.~Zhou, G.~Huang, and T.~Darrell, ``Rethinking the value of network pruning,'' \emph{arXiv preprint arXiv:1810.05270}, Mar. 2018.

\end{thebibliography}
\bibliographystyle{IEEEtran}

\end{document}